\batchmode
\makeatletter
\def\input@path{{"C:/Trabajo laptop/Mis articulos/Finished/General SDE PMBM filter/Accepted/"}}
\makeatother
\documentclass[onecolumn,british,english,twocolumn]{IEEEtran}
\usepackage[T1]{fontenc}
\usepackage[latin9]{inputenc}
\usepackage{amsmath}
\usepackage{amsthm}
\usepackage{amssymb}
\usepackage{stmaryrd}
\usepackage{graphicx}
\PassOptionsToPackage{normalem}{ulem}
\usepackage{ulem}

\makeatletter

\providecommand{\tabularnewline}{\\}

\theoremstyle{plain}
\newtheorem{thm}{\protect\theoremname}
\theoremstyle{plain}
\newtheorem{prop}[thm]{\protect\propositionname}
\theoremstyle{definition}
\newtheorem{example}[thm]{\protect\examplename}
\theoremstyle{plain}
\newtheorem{lem}[thm]{\protect\lemmaname}

\usepackage{cite}
\allowdisplaybreaks 

\usepackage[margin=8pt,font=footnotesize]{caption}
\usepackage{algorithm}
\usepackage{algpseudocode}
\usepackage{amsmath}  

\makeatother

\usepackage{babel}
\addto\captionsbritish{\renewcommand{\examplename}{Example}}
\addto\captionsbritish{\renewcommand{\lemmaname}{Lemma}}
\addto\captionsbritish{\renewcommand{\propositionname}{Proposition}}
\addto\captionsbritish{\renewcommand{\theoremname}{Theorem}}
\addto\captionsenglish{\renewcommand{\examplename}{Example}}
\addto\captionsenglish{\renewcommand{\lemmaname}{Lemma}}
\addto\captionsenglish{\renewcommand{\propositionname}{Proposition}}
\addto\captionsenglish{\renewcommand{\theoremname}{Theorem}}
\providecommand{\examplename}{Example}
\providecommand{\lemmaname}{Lemma}
\providecommand{\propositionname}{Proposition}
\providecommand{\theoremname}{Theorem}

\begin{document}
\title{Gaussian multi-target filtering with target dynamics driven by a stochastic
differential equation}
\author{Ángel F. García-Fernández, Simo Särkkä \IEEEmembership{Senior Member, IEEE}\thanks{A. F. Garc\'ia-Fern\'andez is with ETSI de Telecomunicaci\'on, Universidad Polit\'ecnica de Madrid, 28040 Madrid, Spain (email: angel.garcia.fernandez@upm.es).} 

\thanks{S. Särkkä is with the  Department of Electrical Engineering and Automation, Aalto University, 02150 Espoo, Finland (email: simo.sarkka@aalto.fi).} }
\maketitle
\begin{abstract}
This paper proposes multi-target filtering algorithms in which target
dynamics are given in continuous time and measurements are obtained
at discrete time instants. In particular, targets appear according
to a Poisson point process (PPP) in time with a given Gaussian spatial
distribution, targets move according to a general time-invariant linear
stochastic differential equation, and the life span of each target
is modelled with an exponential distribution. For this multi-target
dynamic model, we derive the distribution of the set of new born targets
and calculate closed-form expressions for the best fitting mean and
covariance of each target at its time of birth by minimising the Kullback-Leibler
divergence via moment matching. This yields a novel Gaussian continuous-discrete
Poisson multi-Bernoulli mixture (PMBM) filter, and its approximations
based on Poisson multi-Bernoulli and probability hypothesis density
filtering. These continuous-discrete  multi-target filters are also
extended to target dynamics driven by nonlinear stochastic differential
equations. 
\end{abstract}


\section{Introduction}

Multi-target filtering consists of estimating the current set of targets
given a sequence of past measurements \cite{Challa_book11,Meyer18,Houssineau18}.
It is a fundamental component in many applications including space-object
surveillance \cite{Delande19}, self-driving vehicles \cite{Pang21}
and maritime collision avoidance systems \cite{Kufoalor20}. Target
dynamic models usually arise from physics, whose laws are in continuous
time, while sensors take measurements at some discrete time steps
\cite{Lambert22,Sarkka_book19}. In addition, continuous-time dynamic
models are decoupled from the sensor sampling times providing a mathematically
principled model in scenarios where sensors provide measurements with
non-uniform time stamps, for instance, asynchronous multi-rate sensors
\cite{Hu10}, non-uniform sampling radars \cite{Eng_thesis07,Zhang06},
event-triggered sensing \cite{Ge20c}, and the automatic identification
system \cite{Uney19}. Therefore, it is important to develop continuous-discrete
multi-target filtering algorithms, in which the multi-target dynamics
are modelled in continuous time and sensor measurements are provided
at some discretised time steps. 

Target dynamics in continuous time are usually modelled via stochastic
differential equations (SDEs) \cite{Oksendal_book03,Sarkka_book19}.
Some widely-used dynamic models driven by SDEs are, for example, the
Wiener velocity model \cite{Sarkka_book19}, the Wiener acceleration
model \cite{Li03}, the variants of the Ornstein-Uhlenbeck (OU) process
\cite{Millefiori16,Millefiori16b,Coraluppi12}, the coordinated turn
model \cite{Li03,Morelande05b} and the re-entry tracking model \cite{Tronarp19}.
In continuous-discrete single-target filtering, we can compute the
posterior density of the target by the standard Bayesian filtering
recursion, which consists of prediction and update. In this case,
the transition density between two time steps can be calculated using
the Fokker-Planck-Kolmogov forward partial differential equation. 

With linear and time-invariant SDEs, the transition density is Gaussian
and its parameters can be computed via the matrix fraction decomposition
\cite{Sarkka_book19,Axelsson15,Van_Loan78}. With non-linear SDEs,
the transition density is non-Gaussian. In this case, Gaussian assumed
density filters can be applied by discretising the SDE first, and
then applying a non-linear Gaussian filter prediction step, e.g.,
an extended or sigma-point Kalman filter \cite{Sarkka12b,Arasaratnam10}.
Another alternative is to use assumed density filtering and propagate
the mean and covariance matrix through the associated ordinary differential
equation (ODE) using analytical linearisation or sigma-points to approximate
the required expectations \cite{Sarkka_book19,Sarkka12b}. This approach
can be derived from the minimisation of the variational Kullback-Leibler
divergence (KLD) \cite{Lambert22}.

Apart from target dynamics modelled by SDEs, a continuous-time model
for multi-target systems must include continuous time models for target
appearances and disappearances. As in \cite{Angel20,Angel21_d,Coraluppi14},
we use an $\mathrm{M}/\mathrm{M}/\infty$ queuing system \cite{Kleinrock_book76},
in which the time of appearance of the targets is a Poisson process,
and their life spans are exponentially distributed. The resulting
discretised multi-target system is Markovian, with the properties
that the target birth model is a Poisson point process (PPP), targets
move independently according to the SDE, and targets remain in the
scene with a certain probability of survival \cite{Angel20}. If this
multi-target system is observed with a standard point-target measurement
model, the posterior density of the set of targets is a Poisson multi-Bernoulli
mixture (PMBM) \cite{Williams15b,Angel18_b}. 

The PMBM filter can be considered a state-of-the-art, fully Bayesian
multiple hypothesis tracking algorithm \cite{Brekke18}, and its multi-target
dynamic model considers discrete-time single-target transition densities,
a probability of survival at each time step, and a PPP for new born
targets at each time step. The PMBM posterior is the union of two
independent processes: a PPP that contains information on targets
that remain undetected, and a multi-Bernoulli mixture that contains
information on potential targets detected at some point up to the
current time step. The PPP therefore is very useful in applications
where information on occluded objects is important, such as self-driving
vehicles and search-and-track applications \cite{Bostrom-Rost21}. 

To develop a Gaussian implementation of the continuous-discrete PMBM
(CD-PMBM) filter, it is required to approximate the single-target
birth density of the PPP for new born targets as a Gaussian and to
use the Gaussian prediction step based on the SDE for the single-target
densities of surviving targets. Closed-form formulas to obtain an
optimal Gaussian fit to the single-target birth density that minimises
the KLD for the Wiener velocity model were provided in \cite{Angel20}. 

While the Wiener velocity model is widely-used in tracking due to
its simplicity, it is usually limited to short-term predictions, and
is inaccurate in long-term predictions, for instance, due to a possibly
unbounded, unrealistic velocity. This limitation can be solved using
OU processes, employed in maritime traffic tracking \cite{Millefiori16,Millefiori16b},
or via more accurate physics-based models such as in space-object
tracking \cite{Horwood11} and projectile tracking \cite{Crouse15}.
Therefore, the main goal of this paper is the development of the Gaussian
CD-PMBM filters for single-target dynamics driven by linear and non-linear
SDEs. The main challenge to develop these multi-target filters is
the calculation of the birth process at the discretised time steps.
 This development enables us to solve multi-target filtering problems
with continuous-time dynamics from first mathematical principles.

In this paper, we make these contributions to enable principled multi-target
filters with SDE dynamics: 1) Calculation of the closed-form optimal
Gaussian approximation of the single-target birth density that minimises
the KLD for linear SDEs, without resorting to sampling; 2) Calculation
of the single-target birth density when the targets appear with the
steady-state solution; 3) Implementation of the Gaussian CD-PMBM filter
and related multi-target filters such as the Poisson multi-Bernoulli
(PMB) filter \cite{Williams15b}, the probability hypothesis density
(PHD) filter \cite{Mahler_book14} and the cardinality probability
hypothesis density (CPHD) filter \cite{Mahler_book14}; 4) The extension
of these filters when target dynamics are driven by non-linear SDEs.

The rest of the paper is organised as follows. The problem formulation
and background are provided in Section \ref{sec:Problem-formulation}.
The discretised multi-target model is given in Section \ref{sec:Discretised-multi-target-dynamic}.
Section \ref{sec:Continuous-discrete-Gaussian-filters} explains the
continuous-discrete Gaussian multi-target filters, and also the extension
to non-linear SDEs. Simulation results are analysed in Section \ref{sec:Simulation-results}.
Finally, conclusions are drawn in Section \ref{sec:Conclusions}.

\section{Problem formulation and background\label{sec:Problem-formulation}}

In this paper, we are interested in approximating the density of the
current set of targets given the sequence of the set of measurements
up to the current time step when multi-target dynamics are given in
continuous time. The dynamic and measurement models are presented
in Section \ref{subsec:Models}. An overview of the solution in Section
\ref{subsec:Overview-solution}. We also provide the required background
on integral computations involving matrix exponentials in Section
\ref{subsec:Integral-computations}.

\subsection{Multi-target dynamic and measurement models\label{subsec:Models}}

We consider a multi-target dynamic model in which target appearance,
dynamic and disappearance models are in continuous time. The set of
targets at a time $t\geq0$ is denoted by $X(t)\in\mathcal{F}\left(\mathbb{R}^{n_{x}}\right)$,
where $\mathbb{R}^{n_{x}}$ is the single target space, and $\mathcal{F}\left(\mathbb{R}^{n_{x}}\right)$
is the space of all finite subsets of $\mathbb{R}^{n_{x}}$ \cite{Mahler_book14}.
Newly appearing targets can be added to $X(t)$ at any $t$, and disappearing
targets can also be removed from $X(t)$ at any time. We consider
the continuous-time multi-target model in \cite{Angel20}, with the
assumptions:
\begin{itemize}
\item A1 Target appearance times are distributed according to a Poisson
process (in time) with rate $\lambda>0$ \cite{Kleinrock_book76}. 
\item A2 At the time a target appears, its distribution is Gaussian with
mean $\overline{x}_{a}$ and covariance matrix $P_{a}$, and is independent
of the rest of the targets. 
\item A3 The life span of a target, the time from its appearance to its
disappearance, follows an exponential distribution with rate $\mu>0$
that is independent of any other variable.
\item A4 Targets move independently according to a linear time-invariant
SDE \cite[Eq. (6.1)]{Sarkka_book19}
\begin{align}
dx\left(t\right) & =Ax\left(t\right)dt+udt+Ld\beta\left(t\right),\label{eq:target_SDE}
\end{align}
where $A\in\mathbb{R}^{n_{x}\times n_{x}}$ and $L\in\mathbb{R}^{n_{x}\times n_{\beta}}$
are matrices, $dx\left(t\right)$ is the differential of $x\left(t\right)$,
$u\in\mathbb{R}^{n_{x}}$, and $\beta\left(t\right)\in\mathbb{R}^{n_{\beta}}$
is a Brownian motion with diffusion matrix $Q_{\beta}$. 
\end{itemize}
A sensor obtains measurements from the set of targets $X(t)$ at
known time instants $t_{k}$, $k\in\mathbb{N}$. The set of targets
at time step $k$, corresponding to time $t_{k}$, is $X_{k}=X\left(t_{k}\right)$
and the set of measurements is $Z_{k}\in\mathcal{F}\left(\mathbb{R}^{n_{z}}\right)$,
where $\mathbb{R}^{n_{z}}$ is the single-measurement space. We consider
the standard point-target detection model \cite{Mahler_book14} such
that, given $X_{k}$, each target $x\in X_{k}$ is detected with probability
$p^{D}\left(x\right)$ generating a measurement with conditional density
$l\left(\cdot|x\right)$, or missed with probability $1-p^{D}\left(x\right)$.
The set $Z_{k}$ contains these target-generated measurements and
clutter measurements, which are distributed according to a PPP in
$\mathbb{R}^{n_{z}}$ with intensity $\lambda^{C}\left(\cdot\right)$. 

\subsection{Overview of the solution\label{subsec:Overview-solution}}

In this paper, we discretise the continuous-time dynamic model in
the previous section at the time steps that the measurements are received.
This results in a standard multi-target dynamic model with a time-varying
probability of survival, single-target transition density and a PPP
birth process \cite{Angel20}. Together with the measurement model,
this implies that the density $f_{k|k'}\left(\cdot\right)$ of $X_{k}$
given the sequence of measurements $\left(Z_{1},...,Z_{k'}\right)$,
with $k'\in\left\{ k-1,k\right\} $, is a PMBM \cite{Williams15b,Angel18_b}. 

The PMBM is the union of two independent processes: a PPP with density
$f_{k|k'}^{\mathrm{p}}\left(\cdot\right)$, representing targets that
have not been detected yet, and a multi-Bernoulli mixture (MBM) with
density $f_{k|k'}^{\mathrm{mbm}}\left(\cdot\right)$, representing
targets that have been detected at some point in the past. The PMBM
density is
\begin{align}
f_{k|k'}\left(X_{k}\right) & =\sum_{Y\uplus W=X_{k}}f_{k|k'}^{\mathrm{p}}\left(Y\right)f_{k|k'}^{\mathrm{mbm}}\left(W\right),\label{eq:PMBM}\\
f_{k|k'}^{\mathrm{p}}\left(X_{k}\right) & =e^{-\int\lambda_{k|k'}\left(x\right)dx}\prod_{x\in X_{k}}\lambda_{k|k'}\left(x\right),\\
f_{k|k'}^{\mathrm{mbm}}\left(X_{k}\right) & =\sum_{a\in\mathcal{A}_{k|k'}}w_{k|k'}^{a}\sum_{\uplus_{l=1}^{n_{k|k'}}X^{l}=X_{k}}\prod_{i=1}^{n_{k|k'}}f_{k|k'}^{i,a^{i}}\left(X^{i}\right),\label{eq:MBM}
\end{align}
where $\lambda_{k|k'}\left(\cdot\right)$ is the PPP intensity, $\mathcal{A}_{k|k'}$
is the set of global hypotheses, $w_{k|k'}^{a}$ is the weight of
global hypothesis $a=\left(a^{1},...,a^{n_{k|k'}}\right)\in\mathcal{A}_{k|k'}$,
$n_{k|k'}$ is the number of Bernoulli components, $a^{i}$ is the
index for the local hypotheses of the $i$-th Bernoulli component
and $f_{k|k'}^{i,a^{i}}\left(\cdot\right)$ is the density of the
$i$-th Bernoulli component with local hypothesis $a^{i}$. Index
$a^{i}\in\{1,...,h_{k|k'}^{i}\}$, where $h_{k'|k}^{i}$ is the number
of local hypotheses for the $i$-th Bernoulli. The symbol $\uplus$
represents the disjoint union and the sum in (\ref{eq:PMBM}) is taken
over all mutually disjoint (and possibly empty) sets $Y$ and $W$
whose union is $X_{k}$. A complete explanation of the PMBM density
as well as the PMBM filtering recursion for a discrete-time multi-target
dynamic model can be found in \cite{Williams15b,Angel18_b}. 

In this paper, we will develop the Gaussian implementation of the
CD-PMBM filter by discretising the multi-target dynamic model resulting
from Assumptions A1-A4. Apart from the CD-PMBM filter, we also develop
other approximate filters with lower computational burden. For example,
the (track-oriented) PMB filter approximates the posterior as a PMB
by performing KLD minimisation after each update using auxiliary variables
\cite{Williams15b,Angel20_e}. The PHD filter and CPHD filters are
obtained by approximating the posterior after each update step as
a PPP and an independent identically distributed cluster process,
respectively, by minimising the KLD \cite{Mahler_book14,Williams15,Angel15_d}. 

\subsection{Integral computations\label{subsec:Integral-computations}}

To develop the Gaussian implementation of the CD-PMBM filter, we will
require to calculate the following types of integrals involving matrix
exponentials \cite{Van_Loan78,Carbonell08}. The identity matrix of
size $p$ is denoted by $I_{p}$. When $p=n_{x}$, we just use the
notation $I$. The zero-matrix of size $m\times n$ is denoted by
$0_{m,n}$. 

\subsubsection{Integral of type 1}

For any $n_{x}\times n_{x}$ matrix $A$ and $n_{x}\times p$ matrix
$B$, we have \cite{Van_Loan78}

\begin{align}
H(t) & =\int_{0}^{t}\exp\left(A\tau\right)Bd\tau\nonumber \\
 & =\left[\begin{array}{cc}
I & 0_{n_{x},p}\end{array}\right]\exp\left(H_{c}^{1}t\right)\left[\begin{array}{c}
0_{n_{x},p}\\
I_{p}
\end{array}\right],\label{eq:Integral_type1}
\end{align}
where
\begin{align}
H_{c}^{1} & =\left[\begin{array}{cc}
A & B\\
0_{p,n_{x}} & 0_{p,p}
\end{array}\right].\label{eq:A_bar}
\end{align}
Matrix exponentials can be computed using standard mathematical software.
In addition, the matrix multiplication on the left and the right of
the matrix exponential in (\ref{eq:Integral_type1}) just select the
top right submatrix of $\exp\left(H_{c}^{1}t\right)$ of dimensions
$n_{x}\times p$. Therefore, in the implementation, rather than performing
these matrix multiplications, we can just select the corresponding
submatrix.

\subsubsection{Integral of type 2}

Given the $n_{x}\times n_{x}$ matrices $A$ and $Q_{c}$, the following
equality holds \cite{Van_Loan78} 
\begin{align}
\int_{0}^{t}\exp\left(A\tau\right)Q_{c}\exp\left(A^{T}\tau\right)d\tau & =F_{2}^{T}\left(t\right)G_{1}(t),\label{eq:Integral_type2}
\end{align}
where
\begin{align}
\exp\left(H_{c}^{2}t\right) & =\left[\begin{array}{cc}
F_{1}(t) & G_{1}(t)\\
0 & F_{2}(t)
\end{array}\right],\\
H_{c}^{2} & =\left[\begin{array}{cc}
-A & Q_{c}\\
0 & A^{T}
\end{array}\right].
\end{align}

\subsubsection{Integral of type 3}

Given the $n_{x}\times n_{x}$ matrices $B$, $Q_{c}$, and $A$,
the following equality holds
\begin{align}
\int_{0}^{t}\exp\left(B\tau\right)Q_{c}\left[\int_{0}^{\tau}\exp\left(Ar\right)dr\right]d\tau & =\exp\left(Bt\right)H_{3}(t),\label{eq:integral_type3}
\end{align}
where
\begin{align}
\exp\left(H_{c}^{3}t\right) & =\left[\begin{array}{ccc}
F_{3}(t) & G_{3}(t) & H_{3}(t)\\
0 & F_{4}(t) & G_{4}(t)\\
0 & 0 & F_{5}(t)
\end{array}\right],\label{eq:expm_int_type2}
\end{align}
and
\begin{align}
H_{c}^{3} & =\left[\begin{array}{ccc}
-B & Q_{c} & 0\\
0 & 0 & I\\
0 & 0 & A
\end{array}\right].\label{eq:H_c3}
\end{align}
Equation (\ref{eq:integral_type3}) is proved in Appendix \ref{sec:Appendix_A}.

\section{Discretised multi-target dynamic model\label{sec:Discretised-multi-target-dynamic}}

This section explains the discretisation of the continuous-time multi-target
model explained in Section \ref{sec:Problem-formulation}. In particular,
for each time step, we need to calculate the probability of survival
(Section \ref{subsec:Probability-of-survival}), the single-target
transition density (Section \ref{subsec:Single-target-transition-density})
and the birth process (Section \ref{subsec:Birth-process}). The birth
process when the mean and covariance at appearance time are given
by the steady-state solution of the SDE is provided in Section \ref{subsec:Steady-state-birth}.
We use the notation $\Delta t_{k}=t_{k}-t_{k-1}$ to denote the time
difference between time steps $k$ and $k-1$.

\subsection{Probability of survival\label{subsec:Probability-of-survival}}

The probability that a target that was alive at time step $t_{k-1}$
is alive at time step $t_{k}$ is the probability $p_{k}^{S}$ of
survival at time step $k$. Using Assumption A3, its value is \cite{Angel20}
\begin{align}
p_{k}^{S} & =e^{-\mu\Delta t_{k}}.\label{eq:Probability_survival}
\end{align}

\subsection{Single-target transition density\label{subsec:Single-target-transition-density}}

For an SDE of the form (\ref{eq:target_SDE}), the transition density
from time $t_{k-1}$ to $t_{k}$ is a Gaussian of the form \cite[Sec. 6.2]{Sarkka_book19}
\begin{align}
g_{k}\left(x\left(t_{k}\right)\left|x\left(t_{k-1}\right)\right.\right) & =\mathcal{N}\left(x\left(t_{k}\right);F_{k}x\left(t_{k-1}\right)+b_{k},Q_{k}\right),\label{eq:single_target_transition_density}
\end{align}
where
\begin{align}
F_{k} & =\exp\left(A\Delta t_{k}\right),\label{eq:discretised_F}\\
b_{k} & =\int_{0}^{\Delta t_{k}}\exp\left(A\tau\right)d\tau u,\label{eq:discretised_b}\\
Q_{k} & =\int_{0}^{\Delta t_{k}}\exp\left(A\tau\right)LQ_{\beta}L^{T}\exp\left(A\tau\right)^{T}d\tau.\label{eq:discretised_Q}
\end{align}
It should be noted that (\ref{eq:discretised_b}) and (\ref{eq:discretised_Q})
can be computed using (\ref{eq:Integral_type1}) and (\ref{eq:Integral_type2}),
respectively. 

\subsection{Birth process\label{subsec:Birth-process}}

From time $t_{k-1}$ to $t_{k}$, there are targets that appear and
disappear according to A1 and A3. The new born targets at time step
$k$ are then the targets that appeared at some time between $t_{k-1}$
and $t_{k}$, and remain alive at time $t_{k}$. It turns out that,
under A1-A4, the distribution of the new born targets is a PPP \cite{Angel20},
which we proceed to characterise. 

The cardinality distribution $\rho_{k}^{b}\left(\cdot\right)$ of
the new born targets at time step $k$ is \cite{Angel20}
\begin{align}
\rho_{k}^{b}\left(n\right) & =\mathcal{P}\left(n;\frac{\lambda}{\mu}\left(1-e^{-\mu\Delta t_{k}}\right)\right),\label{eq:Cardinality_new_born_targets}
\end{align}
where $\mathcal{P}\left(\cdot;a\right)$ is a Poisson distribution
with parameter $a$. Equation (\ref{eq:Cardinality_new_born_targets})
corresponds to the transient solution of an $\mathrm{M}/\mathrm{M}/\infty$
queueing system, with arrival rate $\lambda$ and service time parameter
$\mu$ \cite{Kulkarni_book16}. 

If a target appears with a time lag $t$, that is, it appears at a
time $t_{k}-t>t_{k-1}$, its density at the time of birth $t_{k}$
is a Gaussian $p_{k}\left(x_{k}\left|t\right.\right)$ with mean and
covariance \cite[Sec. 6.2]{Sarkka_book19}
\begin{align}
\mathrm{E}\left[x_{k}\left|t\right.\right] & =\exp\left(At\right)\overline{x}_{a}+\int_{0}^{t}\exp\left(A\tau\right)d\tau u,\label{eq:E_x_t}\\
\mathrm{C}\left[x_{k}\left|t\right.\right] & =\exp\left(At\right)P_{a}\exp\left(A^{T}t\right)\nonumber \\
 & \quad+\int_{0}^{t}\exp\left(A\tau\right)LQ_{\beta}L^{T}\exp\left(A^{T}\tau\right)d\tau,\label{eq:C_x_t}
\end{align}
where the integrals are of the form in (\ref{eq:Integral_type1})
and (\ref{eq:Integral_type2}), and $\overline{x}_{a}$ and $P_{a}$
and the mean and covariance matrix at the time of appearance, see
A2.

The distribution of the time lag $t$ of new born targets at time
step $k$ is a truncated exponential with parameter $\mu$ in the
interval $\left[0,\Delta t_{k}\right)$, whose density is \cite{Angel20}
\begin{align}
p_{k}\left(t\right) & =\frac{\mu}{1-e^{-\mu\Delta t_{k}}}e^{-\mu t}\chi_{\left[0,\Delta t_{k}\right)}\left(t\right).\label{eq:density_time_lag}
\end{align}
Therefore, the single target birth density of the PPP for new born
targets is
\begin{align}
p_{k}\left(x_{k}\right) & =\int_{0}^{\Delta t_{k}}p_{k}\left(x_{k}\left|t\right.\right)p_{k}\left(t\right)dt.\label{eq:single_target_birth_density}
\end{align}
Using (\ref{eq:Cardinality_new_born_targets}), the new born target
PPP intensity is 
\begin{align}
\lambda_{k}^{B}\left(x_{k}\right) & =\frac{\lambda}{\mu}\left(1-e^{-\mu\Delta t_{k}}\right)p_{k}\left(x_{k}\right).\label{eq:birth_intensity}
\end{align}

\subsubsection{Best Gaussian KLD fit to the birth process}

In this paper, we pursue a Gaussian implementation of the continuous-discrete
multi-target filters. To do so, we make a Gaussian approximation to
the single target birth density (\ref{eq:single_target_birth_density})
by calculating its mean and covariance matrix. The resulting Gaussian
birth PPP is the best fit to the true birth PPP, with intensity (\ref{eq:birth_intensity}),
from a KLD minimisation perspective \cite[Prop. 1]{Angel20}. The
mean and covariance matrix of the single-target birth density are
provided in the following proposition. 
\begin{prop}
\label{prop:mean_cov_birth}Under Assumptions A1-A4, the mean at the
time of birth of a single target is
\begin{align}
\overline{x}_{b,k} & =\mathrm{E}\left[\mathrm{E}\left[x_{k}\left|t\right.\right]\right]\label{eq:mean_birth_prev}\\
 & =\overline{x}_{b,k,1}+\overline{x}_{b,k,2},\label{eq:mean_birth}
\end{align}
where
\begin{align*}
\overline{x}_{b,k,1} & =\frac{\mu}{1-e^{-\mu\Delta t_{k}}}\left[\begin{array}{cc}
I & 0_{n_{x},1}\end{array}\right]\exp\left(\overline{A}_{b}\Delta t_{k}\right)\left[\begin{array}{c}
0_{n_{x},1}\\
1
\end{array}\right],\\
\overline{x}_{b,k,2} & =\frac{\mu}{1-e^{-\mu\Delta t_{k}}}\left[\begin{array}{cc}
I & 0_{n_{x},2}\end{array}\right]\exp\left(\overline{A}_{b,u}t\right)\left[\begin{array}{c}
0_{n_{x}+1,1}\\
1
\end{array}\right],
\end{align*}
and 
\begin{align}
\overline{A}_{b} & =\left[\begin{array}{cc}
A-\mu I & \overline{x}_{a}\\
0_{1,n_{x}} & 0
\end{array}\right],\label{eq:mean_birth_Ab}\\
\overline{A}_{b,u} & =\left[\begin{array}{ccc}
A-\mu I & u & 0_{n_{x},1}\\
0_{1,n_{x}} & -\mu & 1\\
0_{1,n_{x}+1} & 0 & 0
\end{array}\right].\label{eq:mean_birth_Abu2}
\end{align}

The covariance matrix at the time of birth of a single target is
\begin{align}
P_{b,k} & =\mathrm{C}\left[\mathrm{E}\left[x_{k}\left|t\right.\right]\right]+\mathrm{E}\left[\mathrm{C}\left[x_{k}\left|t\right.\right]\right],\label{eq:cov_birth}
\end{align}
where
\begin{align}
\mathrm{E}\left[\mathrm{C}\left[x_{k}\left|t\right.\right]\right] & =-\frac{1}{1-e^{-\mu\Delta t_{k}}}\left[e^{-\mu\Delta t_{k}}\int_{0}^{\Delta t_{k}}\exp\left(A\tau\right)\right.\nonumber \\
 & \quad\times LQ_{\beta}L^{T}\exp\left(A^{T}\tau\right)d\tau\nonumber \\
 & \quad+\int_{0}^{\Delta t_{k}}\exp\left(\left(A-\mu/2I\right)t\right)C_{b}\nonumber \\
 & \quad\left.\times\exp\left(\left(A-\mu/2I\right)^{T}t\right)dt\right],\label{eq:cov_birth_E_C_x}\\
\mathrm{C}\left[\mathrm{E}\left[x_{k}\left|t\right.\right]\right] & =\Sigma_{xx}+\Sigma_{xu}+\Sigma_{xu}^{T}+\Sigma_{uu}-\overline{x}_{b,k}\overline{x}_{b,k}^{T},\label{eq:cov_birth_C_E_x}
\end{align}
where
\begin{align}
C_{b} & =LQ_{\beta}L^{T}+\mu P_{a},\label{eq:C_b}
\end{align}
and
\begin{align}
\Sigma_{xx} & =\frac{\mu}{1-e^{-\mu\Delta t_{k}}}\int_{0}^{\Delta t_{k}}\exp\left(\left(A-\mu/2I\right)t\right)\overline{x}_{a}\overline{x}_{a}^{T}\nonumber \\
 & \quad\times\exp\left(\left(A-\mu/2I\right)^{T}t\right)dt,\label{eq:Sigma_xx}\\
\Sigma_{xu} & =\frac{\mu}{1-e^{-\mu\Delta t_{k}}}\int_{0}^{\Delta t_{k}}\exp\left(\left(A-\mu I\right)t\right)\overline{x}_{a}u^{T}\nonumber \\
 & \quad\times\left(\int_{0}^{t}\exp\left(A^{T}\tau\right)d\tau\right)dt,\label{eq:Sigma_xu}\\
\Sigma_{uu} & =-\Sigma_{uu,1}+\Sigma_{uu,2}+\Sigma_{uu,2}^{T},\\
\Sigma_{uu,1} & =\frac{e^{-\mu\Delta t_{k}}}{1-e^{-\mu\Delta t_{k}}}\int_{0}^{\Delta t_{k}}\exp\left(A\tau\right)d\tau uu^{T}\nonumber \\
 & \quad\times\int_{0}^{\Delta t_{k}}\exp\left(A^{T}\tau\right)d\tau,\label{eq:Sigma_uu1}\\
\Sigma_{uu,2} & =\frac{1}{1-e^{-\mu\Delta t_{k}}}\int_{0}^{\Delta t_{k}}\exp\left(\left(A-\mu I\right)t\right)uu^{T}\nonumber \\
 & \quad\times\left(\int_{0}^{t}\exp\left(A^{T}\tau\right)d\tau\right)dt.\label{eq:Sigma_uu2}
\end{align}

The integrals (\ref{eq:cov_birth_E_C_x}) and (\ref{eq:Sigma_xx})
can be computed using (\ref{eq:Integral_type2}). The integrals in
(\ref{eq:Sigma_uu1}) can be computed using (\ref{eq:Integral_type1}).
The integrals (\ref{eq:Sigma_xu}) and (\ref{eq:Sigma_uu2}) can be
computed using (\ref{eq:integral_type3}). 
\end{prop}
The proof of Proposition \ref{prop:mean_cov_birth} is provided in
Appendix \ref{sec:Appendix_B}. It should be noted that $\overline{A}_{b}$
and $\overline{A}_{b,u}$ are square matrices of order $n_{x}+1$
and $n_{x}+2$, respectively. The mean $\overline{x}_{b,k}$ and covariance
matrix $P_{b,k}$ depend on the disappearance rate $\mu$ of the targets,
the SDE, and the density of the targets at the time of appearance,
characterised by the mean $\overline{x}_{a}$ and covariance matrix
$P_{a}$. In addition, the intensity of the PPP also depends on the
appearance rate $\lambda$ of new targets, see (\ref{eq:birth_intensity}).
Therefore, the Gaussian birth PPP depends on all the parameters of
the multi-target dynamic system. It should also be noted that the
second term in (\ref{eq:cov_birth_E_C_x}) plus $\Sigma_{xx}$ can
be computed with a single matrix exponential, and the same happens
with $\Sigma_{xu}+\Sigma_{uu,2}$.

\begin{figure}
\begin{centering}
\includegraphics[scale=0.3]{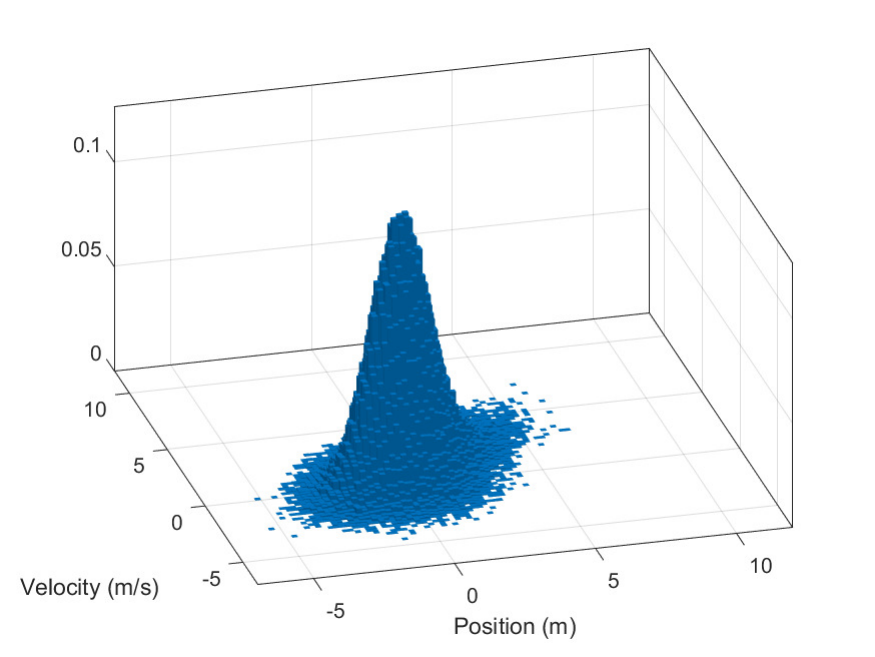}\includegraphics[scale=0.3]{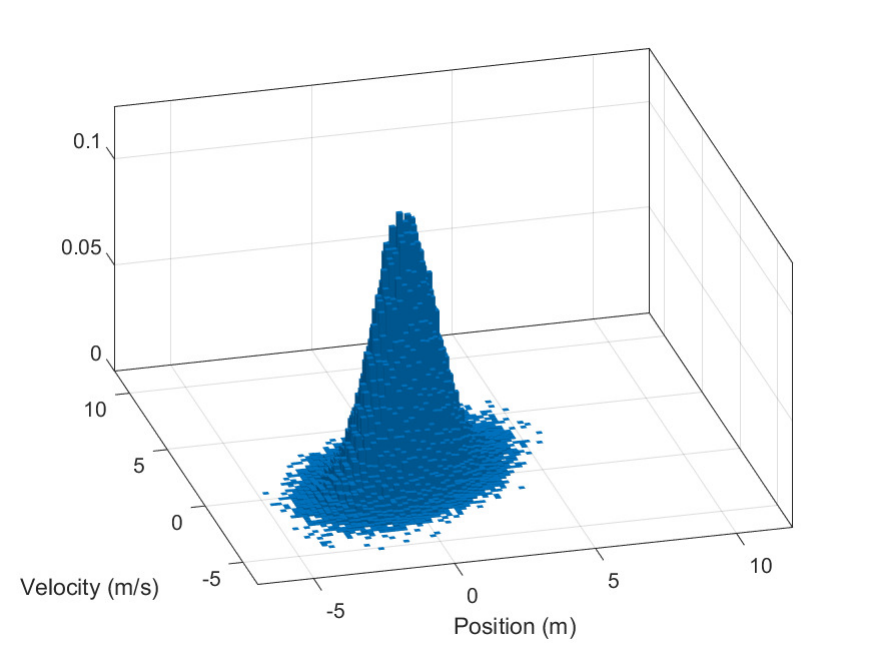}
\par\end{centering}
\begin{centering}
\includegraphics[scale=0.3]{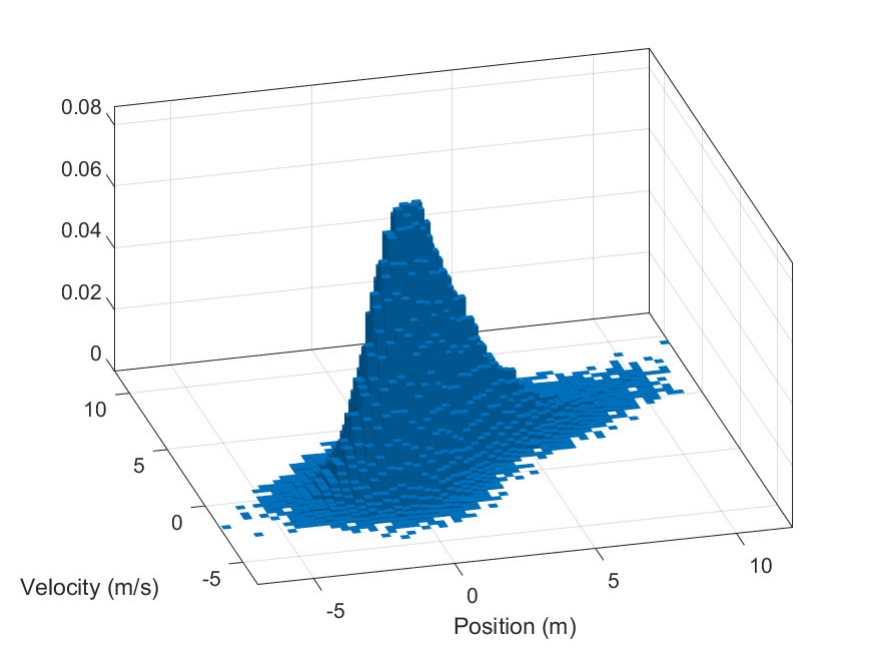}\includegraphics[scale=0.3]{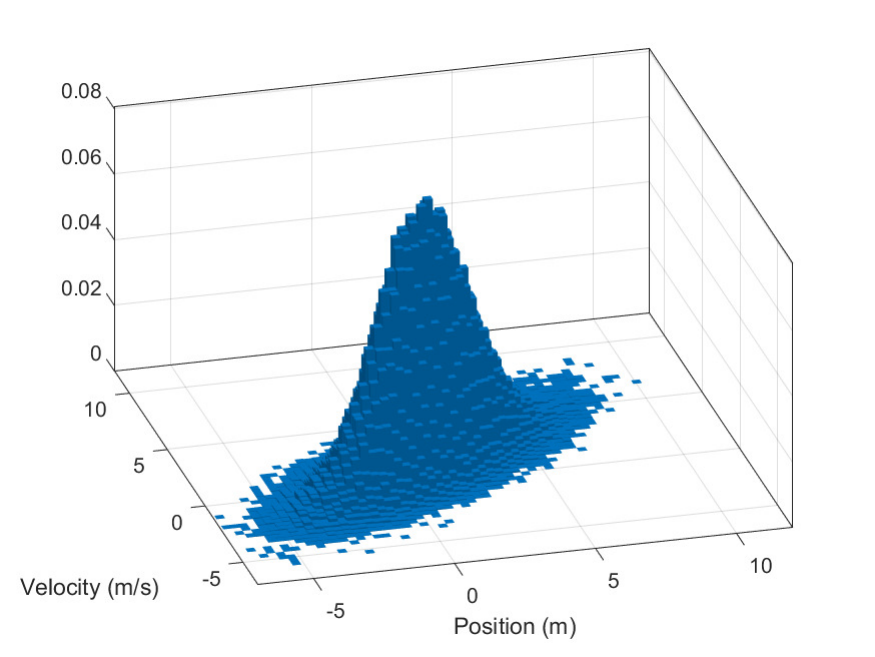}
\par\end{centering}
\caption{\label{fig:Histograms_birth}Normalised histograms of Example \ref{exa:Birth_samples}
with samples from the single target birth density (\ref{eq:single_target_birth_density})
with $\Delta t_{k}=1\,\mathrm{s}$ (top left), from its best Gaussian
fit with $\Delta t_{k}=1\,\mathrm{s}$ (top right), from the single
target birth density with $\Delta t_{k}=2\,\mathrm{s}$ (bottom left),
and its best Gaussian fit $\Delta t_{k}=2\,\mathrm{s}$ (bottom right).}
\end{figure}

\begin{example}
\label{exa:Birth_samples}We consider a one dimensional target with
state $x=\left[p,v\right]^{T}$, where $p$ is its position and $v$
its velocity. Its mean and covariance matrix at appearing time are
$\overline{x}_{a}=\left[0,0\right]^{T}$ and $P_{a}=I_{2}$, with
units in the international system. The target moves with the SDE (\ref{eq:target_SDE})
with
\begin{equation}
A=\left[\begin{array}{cc}
0 & 1\\
0 & -\gamma
\end{array}\right],u=\left[\begin{array}{c}
0\\
\gamma\overline{v}
\end{array}\right],L=\left[\begin{array}{c}
0\\
1
\end{array}\right],
\end{equation}
with $\gamma=0.2$, $\overline{v}=10$, $Q_{\beta}=1$. The velocity
element of this SDE corresponds to a mean-reverting Ornstein-Uhlenbeck
(OU) process, similar to the one in \cite{Millefiori16b}, with long-run
mean velocity $\overline{v}$. The target life span rate is $\mu=0.01$. 

We draw $2\cdot10^{5}$ samples from the single target birth density
(\ref{eq:single_target_birth_density}) first with $\Delta t_{k}=1\,\mathrm{s}$
and then with $\Delta t_{k}=2\,\mathrm{s}$. We obtain the best Gaussian
fit to the single target birth density, using Proposition \ref{prop:mean_cov_birth},
and draw $2\cdot10^{5}$ samples from this process. The resulting
normalised histograms with $\Delta t_{k}=1\,\mathrm{s}$ and $\Delta t_{k}=2\,\mathrm{s}$
are shown in Figure \ref{fig:Histograms_birth}. We can see that,
despite the fact that the mean velocity at the time of appearance
is 0, the mean velocity increases at the time of birth, as the SDE
process tends to have a mean velocity $\overline{v}$. The target
can move a longer distance and gather more speed in $\Delta t_{k}=2\,\mathrm{s}$
than in $\Delta t_{k}=1\,\mathrm{s}$. In addition, for $\Delta t_{k}=1\,\mathrm{s}$,
the best Gaussian fit is closer to the true birth density than with
$\Delta t_{k}=2\,\mathrm{s}$. 

We have also fitted a Gaussian distribution via moment matching using
the samples from (\ref{eq:single_target_birth_density}). The resulting
KLDs between this Gaussian, and the Gaussian obtained using Proposition
\ref{prop:mean_cov_birth} for $\Delta t_{k}=1\,\mathrm{s}$ and $\Delta t_{k}=2\,\mathrm{s}$
are $6.28\cdot10^{-6}$ and $7.03\cdot10^{-6}$, respectively \cite{Hershey07}.
As expected, these KLDs are close to zero. Nevertheless, the results
of Proposition \ref{prop:mean_cov_birth} are obtained at a fraction
of the computational cost of the sampled-based approximation. $\boxempty$
\end{example}

\subsubsection{Birth process via state augmentation}

The mean and covariance matrix of the birth process, given by (\ref{eq:mean_birth})
and (\ref{eq:cov_birth}), can also be computed via state augmentation.
That is, it is possible to write an equivalent SDE to (\ref{eq:target_SDE})
with an augmented state $s(t)$ that stacks $x(t)$ and the non-zero
elements of $u$. The mean and covariance at the time of birth can
then be computed by Proposition \ref{prop:mean_cov_birth} using matrix
exponentials of higher dimensions, but removing the terms associated
with $u$. This approach can be beneficial for low-dimensional states
but becomes less effective as the dimensionality increases, since
computing the matrix exponential of a matrix of size $n$ has a computational
complexity $O(n^{3})$ \cite{Golub_book96}. 

To analyse this, Table \ref{tab:Mean-execution-time} shows the mean
execution times over 1000 Monte Carlo runs for the calculations of
the birth parameters used in Example \ref{exa:Birth_samples} with
variable state dimension. These times were obtained with a Matlab
implementation on an Intel core i5 laptop. We can see that, in this
example, the state augmentation approach is faster to compute the
birth parameters than the standard approach for state dimensions 2
and 4, but it is slower for state dimensions 10 to 40. Nevertheless,
these computational times (for both standard state and augmented state)
do not have any impact on the computational burden of the multi-target
filters in the simulated scenarios, as there are other parts of a
multi-target filter that take longer time, e.g., the data association
problem.

\selectlanguage{british}%
\begin{table}
\selectlanguage{english}%
\caption{\label{tab:Mean-execution-time}Mean execution time (s) to compute
the mean and covariance matrix at time of birth}

\centering{}%
\begin{tabular}{c|cc}
\hline 
State dimension $n_{x}$ &
Standard state &
Augmented state\tabularnewline
\hline 
2 &
$2.54\cdot10^{-4}$ &
$1.74\cdot10^{-4}$\tabularnewline
4 &
$2.65\cdot10^{-4}$ &
$1.89\cdot10^{-4}$\tabularnewline
10 &
$4.99\cdot10^{-4}$ &
$5.59\cdot10^{-4}$\tabularnewline
20 &
0.0012 &
0.0018\tabularnewline
40 &
0.0044 &
0.0100\tabularnewline
\hline 
\end{tabular}\selectlanguage{british}%
\end{table}

\selectlanguage{english}%

\subsection{Steady-state mean and covariance at appearance time\label{subsec:Steady-state-birth}}

The SDE (\ref{eq:target_SDE}) admits a steady-state solution if
the matrix $A$ is Hurwitz, which is a matrix whose eigenvalues have
real parts that are strictly negative \cite[Sec. 6.5]{Khalil_book02,Sarkka_book19}.
In this case, as all the eigenvalues of $A$ are different from zero,
$A$ is invertible and the steady-state mean $\overline{x}_{\infty}$
and covariance $P_{\infty}$ are
\begin{align}
\overline{x}_{\infty} & =-A^{-1}u,\label{eq:mean_steady}\\
P_{\infty} & =\int_{0}^{\infty}\exp\left(A\tau\right)LQ_{\beta}L^{T}\exp\left(A^{T}\tau\right)d\tau,\label{eq:cov_steady}
\end{align}
where (\ref{eq:cov_steady}) is the solution to the Lyapunov equation
\begin{align}
0 & =AP_{\infty}+P_{\infty}A^{T}+LQ_{\beta}L^{T}.
\end{align}
Then, the following lemma holds.
\begin{lem}
\label{lem:Steady_state_appearance}Given a Hurwitz matrix $A$ in
the SDE (\ref{eq:target_SDE}), if the mean and covariance at time
of appearance are the steady-state ones, $\overline{x}_{a}=\overline{x}_{\infty}$
and $P_{a}=P_{\infty}$, given by (\ref{eq:mean_steady}) and (\ref{eq:cov_steady}),
the mean and covariance matrix at the time of birth are the steady-state
ones
\begin{equation}
\overline{x}_{b,k}=\overline{x}_{\infty},\,P_{b,k}=P_{\infty},
\end{equation}
\begin{equation}
\mathrm{E}\left[x_{k}\left|t\right.\right]=\overline{x}_{\infty},\,\mathrm{C}\left[x_{k}\left|t\right.\right]=P_{\infty}.\label{eq:cond_moments_steady}
\end{equation}
\end{lem}
Lemma \ref{lem:Steady_state_appearance} is proved in Appendix \ref{sec:Appendix_C}.
This lemma implies that, under the steady-state appearance conditions,
it is not required to calculate the integrals in Proposition \ref{prop:mean_cov_birth}
to calculate the mean $\overline{x}_{b,k}$ and covariance $P_{b,k}$
at the time of birth, as they correspond to the steady-state ones,
which can be calculated offline. In addition, it should be noted that,
in this case, the birth density is actually Gaussian, while in the
general setting, the birth density is approximated as Gaussian, as
was illustrated in Example \ref{exa:Birth_samples}. An example of
a dynamic model in which $A$ is a Hurwitz matrix is the mixed OU
process in \cite{Coraluppi12} in which both the position and velocity
components include drift terms.

\section{Continuous-discrete Gaussian multi-target filters\label{sec:Continuous-discrete-Gaussian-filters}}

The discretised multi-target dynamic model in the previous section
has a linear/Gaussian single-target transition density (\ref{eq:single_target_transition_density}),
a state-independent probability of survival (\ref{eq:Probability_survival}),
and a PPP birth model with intensity (\ref{eq:birth_intensity}) that
is approximated as Gaussian with mean (\ref{eq:mean_birth}) and covariance
matrix (\ref{eq:cov_birth}). For this multi-target dynamic model
and the measurement model in Section \ref{subsec:Models}, we can
directly apply the PMBM, PMB, PHD and CPHD filters, see Section \ref{subsec:Overview-solution},
giving rise to their continuous-discrete versions, which we refer
to as CD-PMBM, CD-PMB, CD-PHD and CD-CPHD filters. These filters can
also be directly implemented if the target appearance model is a Gaussian
mixture instead of Gaussian by applying Proposition \ref{prop:mean_cov_birth}
to each component \cite[Sec. IV.D and Lem. 6]{Angel20}.

The Gaussian implementations of these filters can then be obtained
by considering a constant probability $p^{D}$ of detection, and a
linear and Gaussian single measurement density $l\left(z|x\right)=\mathcal{N}\left(z;Hx,R\right)$
\cite{Angel18_b,Williams15b,Vo06,Vo07}. The resulting Gaussian CD-PMBM
filtering recursion is provided in Section \ref{subsec:Gaussian-CD-PMBM-filter}.
In Section \ref{subsec:Steady-state-mean-and}, we also explain how
to compute the intensity of the targets that have never been detected
in the CD-PMBM filter when the targets appear with the steady-state
mean and covariance matrix. We also show how to extend these filters
when the SDE is non-linear in Section \ref{subsec:Extension-to-nonlinear}.

\subsection{Gaussian CD-PMBM filter\label{subsec:Gaussian-CD-PMBM-filter}}

In a Gaussian CD-PMBM filter implementation, the PPP intensity is
an unnormalised Gaussian mixture
\begin{align}
\lambda_{k|k'}\left(x\right) & =\sum_{j=1}^{n_{k|k'}^{p}}w_{k|k'}^{p,j}\mathcal{N}\left(x;\overline{x}_{k|k'}^{p,j},P_{k|k'}^{p,j}\right),\label{eq:PPP_intensity_GM}
\end{align}
where $n_{k|k'}^{p}$, is the number of Gaussian components, and $w_{k|k'}^{p,j}$,
$\overline{x}_{k|k'}^{p,j}$ and $P_{k|k'}^{p,j}$ are the weight,
mean and covariance matrix of the $j$-th component. The expected
number of targets that exist at time step $k$ but have not been detected
yet is the sum of the weights $\sum_{j=1}^{n_{k|k'}^{p}}w_{k|k'}^{p,j}$.

In addition, the single-target density of the $i$-th Bernoulli with
local hypothesis $a^{i}$ is
\begin{align}
p_{k|k'}^{i,a^{i}}\left(x\right) & =\mathcal{N}\left(x;\overline{x}_{k|k'}^{i,a^{i}},P_{k|k'}^{i,a^{i}}\right),\label{eq:Single_target_density_Gaussian}
\end{align}
where $\overline{x}_{k|k'}^{i,a^{i}}$ is the mean and $P_{k|k'}^{i,a^{i}}$
is the covariance matrix. 

The CD-PMBM filter is initiated at time $t=0$ ($k=0$), with parameters
$\lambda_{0|0}\left(x\right)=0$, $n_{0|0}^{p}=0$, $n_{0|0}=0$.
The set $\mathcal{A}_{k|k'}$ of global hypotheses and the local data-association
hypotheses are constructed as in the PMBM filter as follows \cite{Williams15b}.
Let $Z_{k}=\left\{ z_{k}^{1},...,z_{k}^{m_{k}}\right\} $ and $\left(k,j\right)$
be a pair to refer to $z_{k}^{j}$. The set of all these pairs up
to time step $k$ is $\mathcal{M}_{k}.$ The local hypothesis $a^{i}$
for the $i$-th Bernoulli considers the associations with the measurements
pairs in $\mathcal{M}_{k}^{i,a^{i}}\subseteq\mathcal{M}_{k}$, such
that in $\mathcal{M}_{k}^{i,a^{i}}$ there can be at maximum one measurement
per time step. The set $\mathcal{A}_{k|k'}$ then contains all possible
$\left(a^{1},...,a^{n_{k|k'}}\right)$ such that $\bigcup_{i=1}^{n_{k|k'}}\mathcal{M}_{k}^{i,a^{i}}=\mathcal{M}_{k}$
and $\mathcal{M}_{k}^{i,a^{i}}\cap\mathcal{M}_{k}^{j,a^{j}}=\emptyset$
$\forall i\neq j$.

The prediction step for the Gaussian CD-PMBM filter is provided in
the following lemma.
\begin{lem}[Prediction]
\label{lem:Prediction_CDPMBM} Let the filtering density at time
step $k-1$ (time $t_{k-1}$) be a PMBM $f_{k-1|k-1}\left(\cdot\right)$
of the form (\ref{eq:PMBM}), with PPP intensity (\ref{eq:PPP_intensity_GM})
and single-target densities (\ref{eq:Single_target_density_Gaussian}).
At time step $k$ (time $t_{k}$), the predicted CD-PMBM filter parameters
are 
\begin{align}
\lambda_{k|k-1}\left(x\right) & =p_{S,k}\sum_{j=1}^{n_{k-1|k-1}^{p}}w_{k-1|k-1}^{p,j}\mathcal{N}\left(x;\overline{x}_{k|k-1}^{p,j},P_{k|k-1}^{p,j}\right)\nonumber \\
 & \quad+\frac{\lambda}{\mu}\left(1-e^{-\mu\Delta t_{k}}\right)\mathcal{N}\left(x;\overline{x}_{b,k},P_{b,k}\right),
\end{align}
where $p_{S,k}$ is given by (\ref{eq:Probability_survival}), $\overline{x}_{b,k}$
and $P_{b,k}$ are given by Proposition \ref{prop:mean_cov_birth},
and 
\begin{align}
\overline{x}_{k|k-1}^{p,j} & =F_{k}\overline{x}_{k-1|k-1}^{p,j}+b_{k},\\
P_{k|k-1}^{p,j} & =F_{k}P_{k-1|k-1}^{p,j}F_{k}^{T}+Q_{k},
\end{align}
where $F_{k}$, $b_{k}$ and $Q_{k}$ are given in (\ref{eq:discretised_F}),
(\ref{eq:discretised_b}) and (\ref{eq:discretised_Q}). The predicted
parameters of the Bernoulli components $i\in\left\{ 1,...,n_{k-1|k-1}\right\} $
are 
\begin{align}
r_{k|k-1}^{i,a^{i}} & =p_{S,k}r_{k-1|k-1}^{i,a^{i}},\\
\overline{x}_{k|k-1}^{i,a^{i}} & =F_{k}\overline{x}_{k-1|k-1}^{i,a^{i}},\\
P_{k|k-1}^{i,a^{i}} & =F_{k}P_{k-1|k-1}^{i,a^{i}}F_{k}^{T}+Q_{k},
\end{align}
and $n_{k|k-1}=n_{k-1|k-1}$, $h_{k|k-1}^{i}=h_{k-1|k-1}^{i}$, $w_{k|k-1}^{i,a^{i}}=w_{k-1|k-1}^{i,a^{i}}$.
\end{lem}
The Gaussian CD-PMBM update is the same as in \cite{Angel20} and
is given in the following lemma.
\begin{lem}[Update]
 \label{lem:Update_CDPMBM}Let the predicted density at time step
$k$ (time $t_{k}$) be a PMBM $f_{k|k-1}\left(\cdot\right)$ of the
form (\ref{eq:PMBM}), PPP intensity (\ref{eq:PPP_intensity_GM})
and single target densities in (\ref{eq:Single_target_density_Gaussian}).
The updated PMBM at time step $k$ with measurement set $Z_{k}=\left\{ z_{k}^{1},...,z_{k}^{m_{k}}\right\} $
is a PMBM of the form (\ref{eq:PMBM}) with PPP intensity $\lambda_{k|k}\left(x\right)=\left(1-p_{D}\right)\lambda_{k|k-1}\left(x\right)$
and $n_{k|k}=n_{k|k-1}+m_{k}$ Bernoulli components. 

For each single target hypothesis of the Bernoulli components at previous
time steps $i\in\left\{ 1,...,n_{k|k-1}\right\} $, the update creates
$\left(m_{k}+1\right)$ single target hypotheses representing a missed
detection and an update with one of the measurements, such that $h_{k|k}^{i}=h_{k|k-1}^{i}\left(m_{k}+1\right)$.
The parameters for missed detection hypotheses, $i\in\left\{ 1,...,n_{k|k-1}\right\} $
$a^{i}\in\left\{ 1,...,h_{k|k-1}^{i}\right\} $, are $\mathcal{M}_{k}^{i,a^{i}}=\mathcal{M}_{k-1}^{i,a^{i}}$,
$\overline{x}_{k|k}^{i,a^{i}}=\overline{x}_{k|k-1}^{i,a^{i}}$, $P_{k|k}^{i,a^{i}}=P_{k|k-1}^{i,a^{i}}.$
\begin{align}
w_{k|k}^{i,a^{i}} & =w_{k|k-1}^{i,a^{i}}\left(1-r_{k|k-1}^{i,a^{i}}+r_{k|k-1}^{i,a^{i}}\left(1-p_{D}\right)\right),\\
r_{k|k}^{i,a^{i}} & =\frac{r_{k|k-1}^{i,a^{i}}\left(1-p_{D}\right)}{1-r_{k|k-1}^{i,a^{i}}+r_{k|k-1}^{i,a^{i}}\left(1-p_{D}\right)}.
\end{align}
For a previous Bernoulli component $i\in\left\{ 1,...,n_{k|k-1}\right\} $
and previous single target hypothesis $\widetilde{a}^{i}\in\left\{ 1,...,h_{k|k-1}^{i}\right\} $,
the new hypothesis generated by measurement $z_{k}^{j}$ has $a^{i}=\widetilde{a}^{i}+h_{k|k-1}^{i}j$,
$r_{k|k}^{i,a^{i}}=1$, and $\mathcal{M}_{k}^{i,a^{i}}=\mathcal{M}_{k-1}^{i,\widetilde{a}^{i}}\cup\left\{ \left(k,j\right)\right\} $
\begin{align}
w_{k|k}^{i,a^{i}} & =w_{k|k-1}^{i,\widetilde{a}^{i}}r_{k|k-1}^{i,\widetilde{a}^{i}}p_{D}\mathcal{N}\left(z_{k}^{j};H\overline{x}_{k|k-1}^{i,\widetilde{a}^{i}},S_{k|k-1}^{i,\widetilde{a}^{i}}\right),\\
\overline{x}_{k|k}^{i,a^{i}} & =\overline{x}_{k|k-1}^{i,\widetilde{a}^{i}}+K_{k|k-1}^{i,\widetilde{a}^{i}}\left(z_{k}^{j}-H\overline{x}_{k|k-1}^{i,\widetilde{a}^{i}}\right),\\
P_{k|k}^{i,a^{i}} & =P_{k|k-1}^{i,\widetilde{a}^{i}}-K_{k|k-1}^{i,\widetilde{a}^{i}}HP_{k|k-1}^{i,\widetilde{a}^{i}},\\
K_{k|k-1}^{i,\widetilde{a}^{i}} & =P_{k|k-1}^{i,\widetilde{a}^{i}}H^{T}\left(S_{k|k-1}^{i,\widetilde{a}^{i}}\right)^{-1},\\
S_{k|k-1}^{i,\widetilde{a}^{i}} & =HP_{k|k-1}^{i,\widetilde{a}^{i}}H^{T}+R.
\end{align}
For the new Bernoulli component created by $z_{k}^{j}$, whose index
is $i=n_{k|k-1}+j$, there are two local hypotheses ($h_{k|k}^{i}=2$).
The first corresponds to a non-existent Bernoulli such that $\mathcal{M}_{k}^{i,1}=\emptyset$,
$w_{k|k}^{i,1}=1$ and $r_{k|k}^{i,1}=0$. The second represents that
the measurement $z_{k}^{j}$ may be clutter or the first detection
of a new target, with parameters $\mathcal{M}_{k}^{i,2}=\left\{ \left(k,j\right)\right\} $
\begin{align}
w_{k|k}^{i,2} & =\lambda^{C}\left(z_{k}^{j}\right)+e\left(z_{k}^{j}\right),\\
e\left(z_{k}^{j}\right) & =p_{D}\sum_{j=1}^{n_{k|k-1}^{p}}w_{k|k-1}^{p,j}\mathcal{N}\left(z_{k}^{j};H\overline{x}_{k|k-1}^{p,j},S_{k|k-1}^{p,j}\right),\\
r_{k|k}^{i,2} & =\frac{e\left(z_{k}^{j}\right)}{\lambda^{C}\left(z_{k}^{j}\right)+e\left(z_{k}^{j}\right)},\\
\overline{x}_{k|k}^{i,2} & =\sum_{l=1}^{n_{k|k-1}^{p}}w_{l}^{j}\overline{x}_{l}^{j},\label{eq:mean_new_Bernoulli}\\
P_{k|k}^{i,2} & =\sum_{l=1}^{n_{k|k-1}^{p}}w_{l}^{j}\left[P_{l}+\left(\overline{x}_{l}^{j}-\overline{x}_{k|k}^{i,2}\right)\left(\overline{x}_{l}^{j}-\overline{x}_{k|k}^{i,2}\right)^{T}\right],\label{eq:cov_new_Bernoulli}\\
w_{l}^{j} & \propto w_{k|k-1}^{p,l}\mathcal{N}\left(z_{k}^{j};H\overline{x}_{k|k-1}^{p,l},S_{k|k-1}^{p,l}\right),\\
\overline{x}_{l}^{j} & =\overline{x}_{k|k-1}^{p,l}+K_{l}\left(z_{k}^{j}-H\overline{x}_{k|k-1}^{p,l}\right),\\
P_{l} & =P_{k|k-1}^{p,l}-K_{l}HP_{k|k-1}^{p,l},\\
K_{l} & =P_{k|k-1}^{p,l}H^{T}\left(S_{k|k-1}^{p,l}\right)^{-1},\\
S_{k|k-1}^{p,l} & =HP_{k|k-1}^{p,l}H^{T}+R.
\end{align}
\end{lem}
In the Gaussian CD-PMBM implementation, the single-target birth density
of the new Bernoulli components is a Gaussian mixture so its mean
and covariance, given by (\ref{eq:mean_new_Bernoulli}) and (\ref{eq:cov_new_Bernoulli}),
are obtained by moment matching \cite{Angel20}. In addition, in a
practical implementation, it is required to prune the number of global
hypothesis \cite[Sec. V.D]{Granstrom20}. This can be achieved using
standard PMBM pruning techniques: gating, Murty's algorithm, discarding
PPP components with low weight, and discarding Bernoulli densities
with low probability of existence \cite{Angel18_b}.

\subsection{Undetected target intensity for steady-state appearance model\label{subsec:Steady-state-mean-and}}

In this section, we calculate the intensity of the targets that have
not been detected yet for the CD-PMBM filter when targets appear with
the steady-mean and covariance matrix at appearance time, see Section
\ref{subsec:Steady-state-birth}, and the measurement model has a
constant probability $p^{D}$ of detection. 
\begin{lem}
\label{lem:Steady_state_mean_cov}If matrix $A$ in the SDE (\ref{eq:target_SDE})
is Hurwitz, the mean and covariance at time of appearance are the
steady-state ones, $\overline{x}_{a}=\overline{x}_{\infty}$ and $P_{a}=P_{\infty}$,
given by (\ref{eq:mean_steady}) and (\ref{eq:cov_steady}), and the
probability of detection is a constant \textup{$p^{D}$}, then, the
intensity of the targets that remain undetected in the CD-PMBM filter,
see (\ref{eq:PMBM}), is
\begin{align}
\lambda_{k|k'}\left(x\right) & =\overline{\lambda}_{k|k'}\mathcal{N}\left(x;\overline{x}_{\infty},P_{\infty}\right),\label{eq:intensity_undetected}
\end{align}
where $\overline{\lambda}_{k|k'}$ can be calculated recursively as
follows
\begin{align}
\overline{\lambda}_{k|k-1} & =p_{k}^{S}\overline{\lambda}_{k-1|k-1}+\frac{\lambda}{\mu}\left(1-e^{-\mu\Delta t_{k}}\right),\label{eq:mean_intensity_prediction}\\
\overline{\lambda}_{k|k} & =\left(1-p^{D}\right)\overline{\lambda}_{k|k-1},\label{eq:mean_intensity_update}
\end{align}
where $\overline{\lambda}_{0|0}$ is the (known) expected number of
targets at time step 0 (usually $\overline{\lambda}_{0|0}=0$) and
the probability of survival is $p_{k}^{S}=e^{-\mu\Delta t_{k}}$,
see (\ref{eq:Probability_survival}). 
\end{lem}
The proof of (\ref{eq:intensity_undetected})-(\ref{eq:mean_intensity_update})
can be directly obtained by applying the PMBM prediction and update
for targets that remain undetected \cite{Williams15b}, and the facts
that the birth intensity is (\ref{eq:birth_intensity}), and the steady-state
mean and covariance do not change in prediction. Therefore, in this
case, the prediction and update of the PPP of the PMBM filter just
requires the calculation of $\overline{\lambda}_{k|k'}$ using (\ref{eq:mean_intensity_prediction})
and (\ref{eq:mean_intensity_update}), as $\overline{x}_{\infty}$
and $P_{\infty}$ are calculated beforehand. In addition, we can
also establish the following lemma for steady-state solutions for
$\overline{\lambda}_{k|k-1}$ and $\overline{\lambda}_{k|k}$.
\begin{lem}
\label{lem:Steady_state_lambda}If the sampling time interval is a
constant $\Delta t$ and the probability of detection is a constant
$p^{D}$, then, the probability of survival is a constant $p^{S}$
given by (\ref{eq:Probability_survival}), and the predicted and updated
expected number of undetected targets ($\overline{\lambda}_{k|k-1}$
and $\overline{\lambda}_{k|k}$) reach steady-state solutions, given
by
\begin{align}
\overline{\lambda}_{k|k-1}^{\infty} & =\frac{\overline{\lambda}^{B}}{1-p^{S}+p^{S}p^{D}},\label{eq:predicted_lambda_steady}\\
\overline{\lambda}_{k|k}^{\infty} & =\frac{\left(1-p^{D}\right)\overline{\lambda}^{B}}{1-p^{S}+p^{S}p^{D}},\label{eq:updated_lambda_steady}
\end{align}
where the (constant) expected number of new born targets at each time
step is $\overline{\lambda}^{B}=\frac{\lambda}{\mu}\left(1-e^{-\mu\Delta t}\right)$, 
\end{lem}
The proof of Lemma \ref{lem:Steady_state_lambda} is provided in Appendix
\ref{sec:Appendix_D}. If the conditions of Lemmas \ref{lem:Steady_state_mean_cov}
and \ref{lem:Steady_state_lambda} hold, and the expected number of
targets at time step 0 (before any measurements are taken) is $\overline{\lambda}_{0|0}=\overline{\lambda}_{k|k}^{\infty}$,
then the expected number of undetected targets is always $\overline{\lambda}_{k|k-1}^{\infty}$
in prediction and $\overline{\lambda}_{k|k}^{\infty}$ in the update,
with an intensity of the form (\ref{eq:intensity_undetected}). This
result also indicates when and how the CD-PMBM filter does not need
to propagate explicit information on undetected targets to perform
tracking on detected targets using first Bayesian principles.\foreignlanguage{british}{ }

It should be noted that a target dynamic model that admits a steady-state
solution is the mixed OU process in \cite{Coraluppi12,Coraluppi14}.
This process can be used to model targets that are likely to move
in a certain area, e.g., patrol agents. In this case, choosing a target
appearance model that matches the steady-state solution simplifies
the CD-PMBM filtering recursion since the PPP intensity is given by
Lemma \ref{lem:Steady_state_mean_cov}, and Lemma \ref{lem:Steady_state_lambda}
for a constant time interval.

\subsection{Extension to nonlinear SDEs\label{subsec:Extension-to-nonlinear}}

In this section, we extend the results in Sections \ref{sec:Discretised-multi-target-dynamic}
and \ref{sec:Continuous-discrete-Gaussian-filters} to the case in
which the targets move with a non-linear time-invariant SDE. We consider
Assumptions A1-A3 and
\begin{itemize}
\item A5 Targets move independently according to a non-linear time-invariant
SDE
\begin{align}
dx\left(t\right) & =f\left(x\right)dt+L\left(x\right)d\beta\left(t\right),\label{eq:nonlinear_SDE}
\end{align}
where $f\left(x\right)$ is the drift function and $L\left(x\right)$
the dispersion matrix.
\end{itemize}
In this case, the update step of the Gaussian CD-PMBM is also given
by Lemma \ref{lem:Update_CDPMBM}, as A5 only affects the dynamic
model. In the prediction step, the probability of survival is also
given by (\ref{eq:Probability_survival}), and has the form of Lemma
\ref{lem:Prediction_CDPMBM}. Under A5, the difference with the previous
results is that the single-target transition density changes. In turn,
this implies that the predicted mean and covariance for surviving
targets are now calculated as in Section \ref{subsec:Prediction-mean-covariance-nonlinear},
and the mean and covariance of a target at the time of birth are calculated
as in Section \ref{subsec:Mean-and-covariance-birth-nonlinear}. These
results are applied in the same manner to the CD-PMB, CD-PHD and CD-CPHD
filters.

\subsubsection{Prediction of the mean and covariance matrix\label{subsec:Prediction-mean-covariance-nonlinear}}

There are several ways to propagate a mean and a covariance matrix
of a target that moves according to the SDE (\ref{eq:nonlinear_SDE})
\cite[Chap. 9]{Sarkka_book19}. Let $m(t)$ and $P(t)$ denote the
mean and covariance of a target that moves according to the SDE (\ref{eq:nonlinear_SDE}).
We use a standard approach based on analytical linearisation of the
SDE. In the following, we drop the dependence on $t$ of $m(t)$ and
$P(t)$ following the standard notation in SDEs \cite{Sarkka_book19}. 

We linearise the drift function $f\left(\cdot\right)$ around the
mean $m$ as
\begin{align}
f\left(x\right) & \approx f\left(m\right)+F_{x}\left(m\right)\left(x-m\right),\label{eq:drift_approx}
\end{align}
where $F_{x}\left(m\right)$ is the Jacobian of $f\left(x\right)$
w.r.t. $x$, and also approximate the dispersion matrix as
\begin{align}
L\left(x\right) & \approx L\left(m\right).\label{eq:dispersion_approx}
\end{align}

Under approximations (\ref{eq:drift_approx}) and (\ref{eq:dispersion_approx}),
the mean and covariance of the target at a given time are obtained
by solving \cite{Sarkka_book19} 
\begin{align}
\frac{dm}{dt} & =f\left(m\right),\label{eq:ODE_linearised_mean}\\
\frac{dP}{dt} & =PF_{x}^{T}\left(m\right)+F_{x}\left(m\right)P+L\left(m\right)QL^{T}\left(m\right).\label{eq:ODE_linearised_cov}
\end{align}
Then, let us consider that the posterior single-target density $p_{k-1|k-1}^{i,a^{i}}\left(\cdot\right)$
is given by (\ref{eq:Single_target_density_Gaussian}). Then, we set
the initial conditions of the ODE (\ref{eq:ODE_linearised_mean})-(\ref{eq:ODE_linearised_cov})
as $m(0)=\overline{x}_{k-1|k-1}^{i,a^{i}}$ and $P(0)=P_{k-1|k-1}^{i,a^{i}}$,
and use an ODE solver to set the predicted mean and covariance as
$\overline{x}_{k|k-1}^{i,a^{i}}=m\left(\Delta t_{k}\right)$ and $P_{k|k-1}^{i,a^{i}}=P\left(\Delta t_{k}\right)$.
This operation is also performed for each Gaussian component in the
PPP (\ref{eq:PPP_intensity_GM}).

\subsubsection{Mean and covariance at the time of birth\label{subsec:Mean-and-covariance-birth-nonlinear}}

Let us consider a target that appears with a time lag $t$. Then,
its conditional mean and covariance at the time of birth are
\begin{align}
\mathrm{E}\left[x_{k}\left|t\right.\right] & =m\left(t\right),\:\mathrm{C}\left[x_{k}\left|t\right.\right]=P\left(t\right),\label{eq:conditional_mean_cov_birth_nonlinear}
\end{align}
where $m\left(t\right)$ and $P\left(t\right)$ are the solution of
the ODE (\ref{eq:ODE_linearised_mean})-(\ref{eq:ODE_linearised_cov})
integrated from time 0 to $t$ with initial condition $m\left(0\right)=\overline{x}_{a}$
and $P\left(0\right)=P_{a}$. Note that the analogous equations for
a linear SDE are given by (\ref{eq:E_x_t}) and (\ref{eq:C_x_t}).
The mean and covariance at the time of birth are then given by the
following proposition.
\begin{prop}
\label{prop:Birth_nonlinear}Under approximations (\ref{eq:drift_approx})
and (\ref{eq:dispersion_approx}) and Assumptions A1-A3 and A5, the
mean $\overline{x}_{b,k}$ and covariance $P_{b,k}$ of a target at
the time of birth are
\begin{align}
\overline{x}_{b,k} & =\frac{\mu}{1-e^{-\mu\Delta t_{k}}}\overline{x}\left(\Delta t_{k}\right),\\
P_{b,k} & =\frac{\mu}{1-e^{-\mu\Delta t_{k}}}\Sigma\left(\Delta t_{k}\right)-\overline{x}_{b,k}\overline{x}_{b,k}^{T},
\end{align}
where $\overline{x}\left(\Delta t_{k}\right)$ and $\Sigma\left(\Delta t_{k}\right)$
are calculated by solving the ODE
\begin{align}
\frac{dm}{dt} & =f\left(m\right),\\
\frac{dP}{dt} & =PF_{x}^{T}\left(m\right)+F_{x}\left(m\right)P+L\left(m\right)QL^{T}\left(m\right),\\
\frac{d\overline{x}}{dt} & =me^{-\mu t},\\
\frac{d\Sigma}{dt} & =\left[P+mm^{T}\right]e^{-\mu t},
\end{align}
from $t=0$ to $\Delta t_{k}$ with initial condition $m\left(0\right)=\overline{x}_{a}$,
$P\left(0\right)=P_{a}$, $\overline{x}\left(0\right)=0$ and $\Sigma\left(0\right)=0$.
\end{prop}
The proof of Proposition \ref{prop:Birth_nonlinear} is provided in
Appendix \ref{sec:Appendix_E}. In practice, these ODEs that involve
mean and covariances can be solved by stacking the mean and covariances
into a single vector and applying an available ODE solver \cite{Sarkka_book19}. 

It should be noted that it is also possible to linearise the drift
function and the dispersion matrix as in (\ref{eq:drift_approx})
and (\ref{eq:dispersion_approx}) at the initial mean $\overline{x}_{a}$,
and leave the linearisation fixed for the whole interval. With the
resulting linearised SDE, the mean and covariance matrix at the time
of birth can be computed via Proposition \ref{prop:mean_cov_birth}.
Nevertheless, since the linearisation errors accumulate over time
as the mean deviates from its initial value, the provided solution
is less accurate than the one provided by Proposition \ref{prop:Birth_nonlinear},
especially for longer time intervals.

\section{Simulation results\label{sec:Simulation-results}}

This section includes the numerical evaluation of the CD-PMBM, CD-PMB,
CD-PHD and CD-CPHD filters in a linear and a non-linear SDE scenario\footnote{Matlab code is available at https://github.com/Agarciafernandez/MTT.}.
In the non-linear scenario, the ODEs are solved using Matlab ode45
solver. The CD-PMBM filter has been implemented with the following
parameters: maximum number of global hypotheses $N_{h}=200$, threshold
for pruning PPP Gaussian components $\Gamma_{p}=10^{-5}$, threshold
for pruning multi-Bernoulli mixture weights $\Gamma_{mbm}=10^{-4}$,
and threshold for the probability of existence to prune Bernoulli
densities $\Gamma_{b}=10^{-5}$. The filter also uses ellipsoidal
gating with threshold 20 and Estimator 1 in \cite{Angel18_b}, with
threshold 0.4. The CD-PMB filter is implemented as the CD-PMBM but
projecting the output of the update step to a PMB density \cite{Williams15b}.
The CD-PHD and CD-CPHD filters have been implemented with a maximum
number of components of 30, pruning threshold $10^{-5}$ and merging
threshold $0.1$. 

We also compare these continuous-discrete filters with the equivalent
(discrete-time) PMBM, PMB, PHD and CPHD filters with a fixed sampling
time, which is taken to be the minimum $\Delta t_{k}$ in the considered
time window. These filters can compute offline the birth model parameters,
which remain fixed for all time steps, using Proposition \ref{prop:mean_cov_birth}
or \ref{prop:Birth_nonlinear}. Then, in the linear case, the equivalent
linear and Gaussian dynamic model can also be computed offline. In
the non-linear case, the propagation of the mean and covariance matrix
of the targets can be done following Section \ref{subsec:Prediction-mean-covariance-nonlinear}.
These filters only perform the update step at the time steps when
measurements have been received. The rest of the parameters of the
PMBM, PMB, PHD and CPHD filters are set as indicated above for the
continuous-discrete versions.

\subsection{Linear SDE\label{subsec:Linear-SDE-simulation} }

We consider an OU process for the velocity as in \cite{Millefiori16b}.
The state is $x\left(t\right)=\left[p_{x}\left(t\right),v_{x}\left(t\right),p_{y}\left(t\right),v_{y}\left(t\right)\right]^{T}$,
which contains position and velocity in the $x$ and $y$ coordinates.
The SDE parameters are
\[
A=\left[\begin{array}{cccc}
0 & 1 & 0 & 0\\
0 & -\gamma & 0 & 0\\
0 & 0 & 0 & 1\\
0 & 0 & 0 & -\gamma
\end{array}\right],u=\left[\begin{array}{c}
0\\
\gamma\overline{v}_{x}\\
0\\
\gamma\overline{v}_{y}
\end{array}\right],L=\left[\begin{array}{cc}
0 & 0\\
1 & 0\\
0 & 0\\
0 & 1
\end{array}\right],
\]
and $Q_{\beta}=qI_{2}$. The numerical values of these parameters
are $\gamma=0.1\,(\mathrm{s}^{-1})$, $\overline{v}_{x}=1\,(\mathrm{m}/\mathrm{s})$,
$\overline{v}_{y}=1\,(\mathrm{m}/\mathrm{s})$, $q=0.2\,(\mathrm{m}^{2}/\mathrm{s}^{3})$.
The rest of the parameters of the multi-target dynamic model are $\lambda=0.08\,\mathrm{s}^{-1}$,
$\mu=0.01\,\mathrm{s}^{-1}$, $\overline{x}_{a}=\left[200\,(\mathrm{m}),3\,(\mathrm{m}/\mathrm{s}),250\,(\mathrm{m}),0\,(\mathrm{m}/\mathrm{s})\right]^{T}$
and $P_{a}=\mathrm{diag}\left(\left[50^{2}\,(\mathrm{m}^{2}),1\,(\mathrm{m^{2}}/\mathrm{s}^{2}),50^{2}\,(\mathrm{m}^{2}),1\,(\mathrm{m}^{2}/\mathrm{s}^{2})\right]\right)$. 

\begin{figure}
\begin{centering}
\includegraphics[scale=0.6]{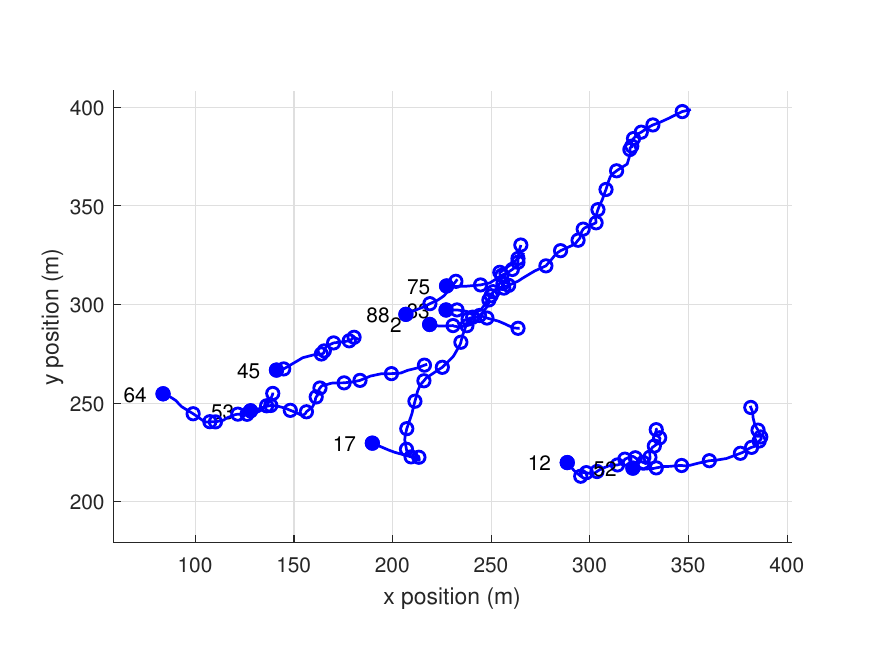}
\par\end{centering}
\caption{\label{fig:Ground-truth-trajectories-linear}Ground truth set of trajectories
with 10 trajectories. Every 10 time steps, the position of a trajectory
is marked with circles. The initial position is marked with a filled
circle, and the number next to it indicates time step of birth.}
\end{figure}

The sensor takes 100 measurements. The time interval $\Delta t_{k}$
between measurements is drawn from an exponential distribution with
parameter $\mu_{m}=1$ \cite{Eng_thesis07}. The resulting time intervals
are the same as in \cite[Fig. 2]{Angel20}. The sensor obtains position
measurements with a probability of detection $p^{D}=0.9$ and
\begin{equation}
H=\begin{bmatrix}\begin{array}{cccc}
1 & 0 & 0 & 0\\
0 & 0 & 1 & 0
\end{array}\end{bmatrix},\,R=\sigma_{r}^{2}I_{2},\label{eq:measurement_model_scenario1}
\end{equation}
where $\sigma_{r}^{2}=4\,\mathrm{m}^{2}$. The clutter intensity is
$\lambda^{C}\left(z\right)=10\cdot u_{A}\left(z\right)$, where $u_{A}\left(\cdot\right)$
is a uniform density in the area $A=\left[0,600\right]\times\left[0,400\right]\,\left(\mathrm{m^{2}}\right)$.
The ground truth set of trajectories, shown in Figure \ref{fig:Ground-truth-trajectories-linear},
is drawn from the above continuous time multi-target model sampled
at the sensor sampling times.

\begin{figure}
\begin{centering}
\includegraphics[scale=0.3]{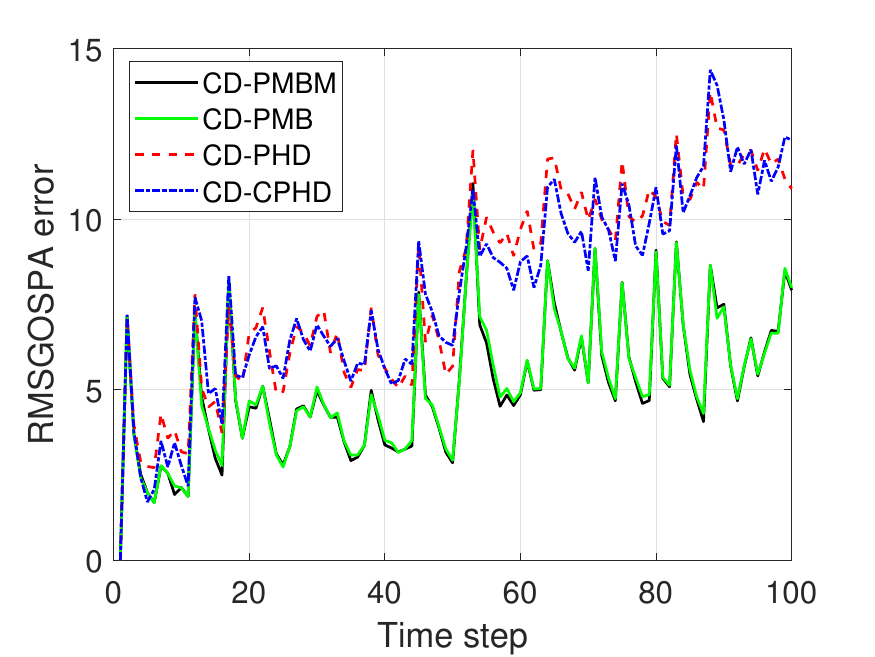}\includegraphics[scale=0.3]{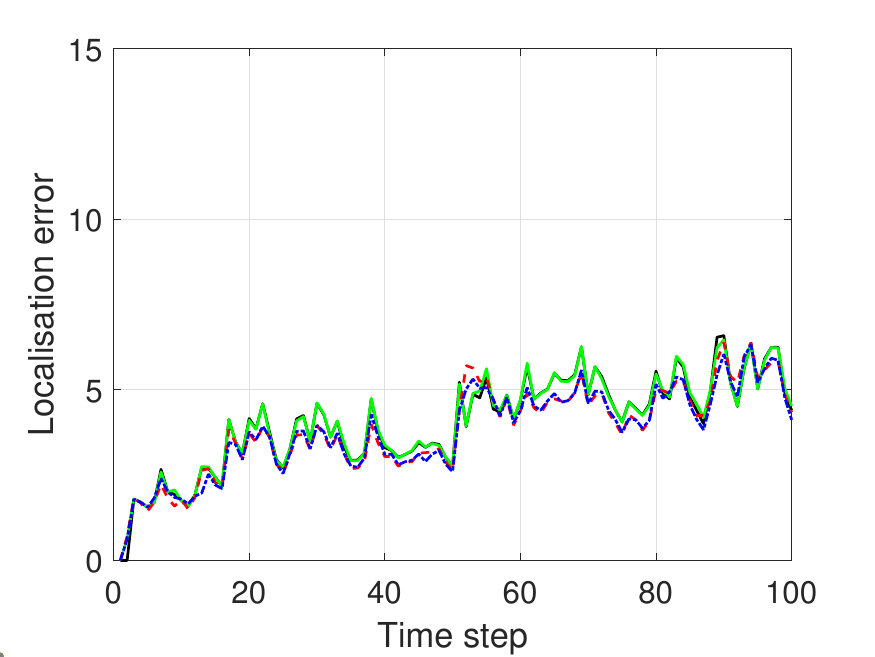}
\par\end{centering}
\begin{centering}
\includegraphics[scale=0.3]{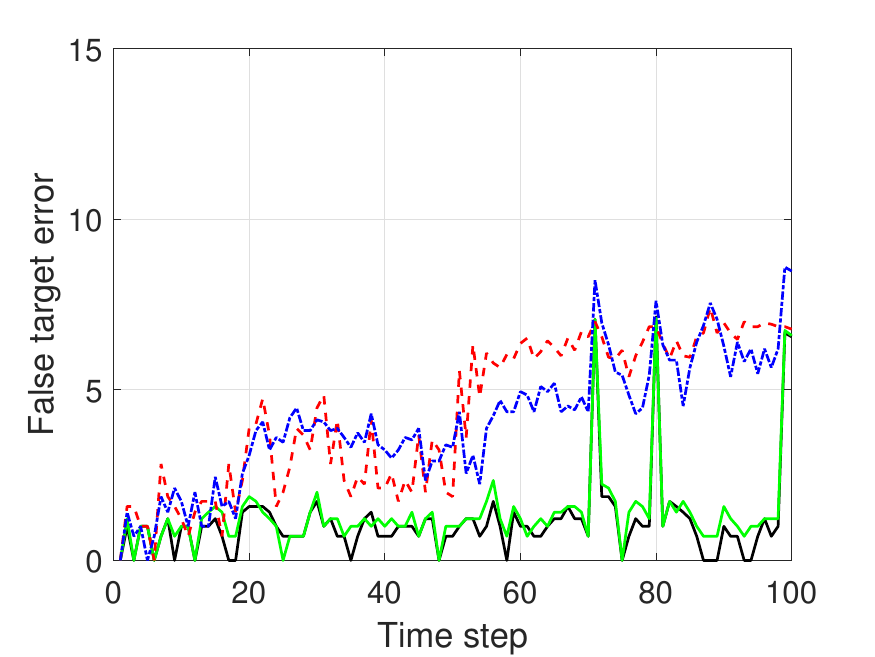}\includegraphics[scale=0.3]{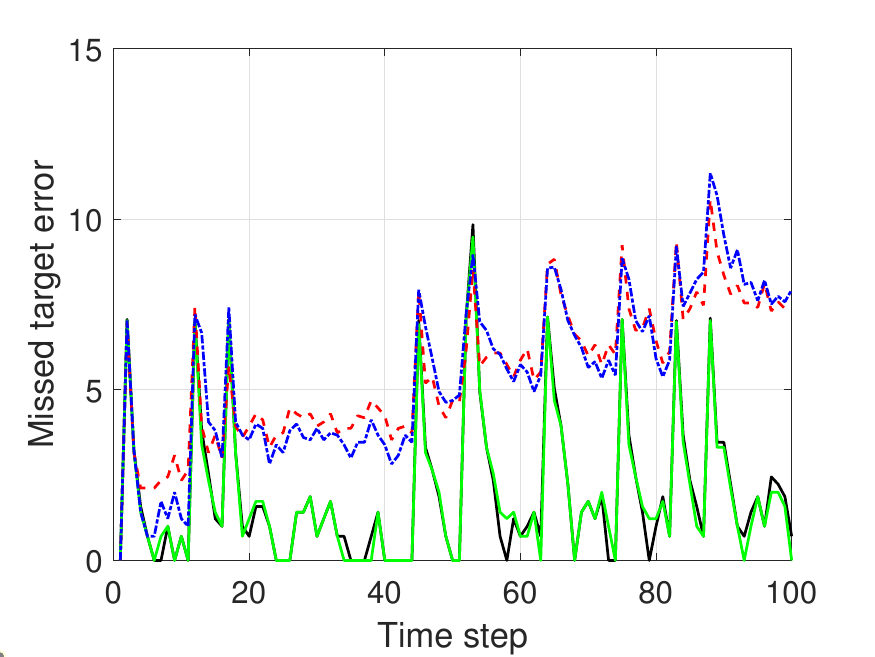}
\par\end{centering}
\caption{\label{fig:RMS-GOSPA-linear}RMS GOSPA positional errors ($\mathrm{m)}$
and their decomposition against time.}
\end{figure}

Filtering performance is evaluated via Monte Carlo simulation with
100 runs. We estimate the set of target positions and calculate its
error using the generalised optimal subpattern assignment (GOSPA)
metric with parameters $\alpha=2$, $c=10\,\mathrm{m}$ and $p=2$
\cite{Rahmathullah17}. The root mean square GOSPA (RMS-GOSPA) at
each time step, and its decomposition in terms of localisation errors,
missed target costs and false target costs for the continuous-discrete
filters are shown in Figure \ref{fig:RMS-GOSPA-linear}. The CD-PMBM
and CD-PMB are the best performing filters. The CD-PHD and CD-CPHD
have worse performance, mainly due to an increase in false target
as well as missed target error. 

The (discrete-time) PMBM, PMB, PHD and CPHD filters, which consider
29374 time steps, produce very similar results to the corresponding
continuous-discrete filters, so the results are not shown in Figure
\ref{fig:RMS-GOSPA-linear}. The difference arises in the computational
time. The computational times to run one Monte Carlo iteration of
the filters on an Intel core i5 laptop are: 5.1 s (CD-PMBM), 1.1 s
(CD-PMB), 1.5 s (CD-PHD), 2.0 s (CD-CPHD), 136.6 s (PMBM), 106.6 s
(PMB), 60.1 s (PHD), 85.5 s (CPHD). We can see that the continuous-discrete
filters show a noteworthy decrease in computational time compared
to the discrete versions (over 26 times faster in the case of the
PMBM). The reason is that the continuous-discrete filters can perform
in a single step the multiple prediction steps that are required in
the discrete versions.

In the following, we also analyse two simpler birth discretisation
methods that approximate the single-target birth density, given by
(\ref{eq:single_target_birth_density}), as a constant across time.
The first method, which we refer to as ``Constant single-target birth
density 1'' (CSBD1), makes the approximation
\begin{align}
p_{k}\left(x_{k}\right) & \approx p_{k}\left(x_{k}\left|\hat{t}\right.\right),\label{eq:approx_birth1}
\end{align}
where $\hat{t}$ is the expected time lag of appearance given the
expected $\Delta t_{k}$, $\mathrm{E}\left[\Delta t_{k}\right]$,
which is 1 second in the simulation. Using \cite[Eq. (36)]{Angel20},
we obtain that
\begin{align}
\hat{t} & =\frac{1}{\mu}-\frac{\mathrm{E}\left[\Delta t_{k}\right]e^{-\mu\mathrm{E}\left[\Delta t_{k}\right]}}{1-e^{-\mu\mathrm{E}\left[\Delta t_{k}\right]}}.
\end{align}
In this scenario, $\hat{t}=0.4992$. The second method, which we refer
to as ``Constant single-target birth density 2'' (CSBD2), neglects
the continuous-time effects by assuming that the single-target birth
density is
\begin{align}
p_{k}\left(x_{k}\right) & \approx\mathcal{N}\left(x;\overline{x}_{a},P_{a}\right).\label{eq:approx_birth2}
\end{align}
In these implementations, the rest of the parameters of the continuous-discrete
filters remain unchanged. 

To notice the effect of different single-target birth discretisations,
the mean and covariance matrix at the time of birth should significantly
differ from those at the time of appearance. With the current appearance
model, experimental results do not show a big difference between these
discretisations. To show differences, we increase the mean velocity
of the appearing targets to $\overline{v}_{a}=[20,-20]^{T}\,(\mathrm{m}/\mathrm{s})$
and $\overline{v}_{a}=[25,-25]^{T}\,(\mathrm{m}/\mathrm{s})$. For
the latter velocity, we also consider an informative birth model with
$P_{a}=I_{4}$, and also a constant $\Delta t_{k}=1\mathrm{s}$. 

\begin{table*}
\caption{\label{tab:RMSGOSPA-linear}RMS-GOSPA error (m) across all time steps
for different single-target birth discretisations in the linear SDE
example}

\centering{}%
\begin{tabular}{c|ccc|ccc|ccc|ccc}
\hline 
 &
\multicolumn{3}{c|}{$\overline{v}_{a}=[20,-20]^{T}$} &
\multicolumn{3}{c|}{$\overline{v}_{a}=[25,-25]^{T}$} &
\multicolumn{3}{c|}{$\overline{v}_{a}=[25,-25]^{T}$,$P_{a}=I_{4}$} &
\multicolumn{3}{c}{$\overline{v}_{a}=[25,-25]^{T}$,$P_{a}=I_{4}$,$\Delta t_{k}=1\mathrm{s}$}\tabularnewline
\hline 
Birth &
Prop. 1 &
CSBD1 &
CSBD2 &
Prop. 1 &
CSBD1 &
CSBD2 &
Prop. 1 &
CSBD1 &
CSBD2 &
Prop. 1 &
CSBD1 &
CSBD2\tabularnewline
\hline 
CD-PMBM &
\uline{5.43} &
5.47 &
5.51 &
\uline{5.43} &
5.47 &
5.54 &
\uline{4.94} &
6.38 &
9.12 &
\uline{4.92} &
5.55 &
5.62\tabularnewline
CD-PMB &
5.46 &
5.48 &
5.54 &
5.46 &
5.50 &
5.53 &
5.04 &
6.24 &
9.14 &
4.95 &
5.57 &
5.60\tabularnewline
CD-PHD &
8.77 &
9.17 &
9.41 &
8.83 &
9.34 &
9.74 &
8.26 &
9.23 &
10.28 &
8.38 &
9.46 &
9.50\tabularnewline
CD-CPHD &
8.45 &
8.80 &
9.14 &
8.48 &
9.12 &
9.61 &
8.16 &
9.37 &
10.75 &
8.26 &
9.61 &
9.64\tabularnewline
\hline 
\end{tabular}
\end{table*}

The RMS-GOSPA errors across all time steps for the different birth
models in the four considered scenarios are shown in Table \ref{tab:RMSGOSPA-linear}.
The birth model of Prop.1 outperforms the other two variants in these
cases. In particular, the improvement in performance for informative
$P_{a}$ is significant for both variable $\Delta t_{k}$ and constant
$\Delta t_{k}$. The reason is that, for informative $P_{a}$, there
are important difference between the single-target birth density at
the time of birth and at the time of appearance, and this has an impact
in filtering performance.

\subsection{Non-linear SDE}

We consider a re-entry tracking problem \cite{Julier04,Tronarp19}.
The state is $x\left(t\right)=\left[p_{x}\left(t\right),v_{x}\left(t\right),p_{y}\left(t\right),v_{y}\left(t\right)\right]^{T}$
and the SDE parameters are

\begin{align}
f\left(x\right) & =I_{2}\otimes\left[\begin{array}{cc}
0 & 1\\
g(x) & d(x)
\end{array}\right]x,\:L=\left[\begin{array}{c}
1\\
1
\end{array}\right]\otimes\left[\begin{array}{cc}
0 & 0\\
1 & 0
\end{array}\right],
\end{align}
where
\begin{align}
g(x) & =-\frac{Gm_{0}}{\left((p_{x})^{2}+(p_{y})^{2}\right)^{3/2}},\\
d(x) & =-\beta_{0}\exp\left(\psi+\frac{r_{0}-\left((p_{x})^{2}+(p_{y})^{2}\right)^{1/2}}{h_{0}}\right)\nonumber \\
 & \quad\times\sqrt{(v_{x})^{2}+(v_{y})^{2}},
\end{align}
where $Gm_{0}=3.986\cdot10^{5}\,\mathrm{km^{3}/s^{2}}$, $\beta_{0}=0.6\,\mathrm{km^{-1}}$,
$\psi=0.7$ is an aerodynamic parameter, $r_{0}=6374\,\mathrm{km},$
$h_{0}=13.4\,\mathrm{km}$ where $Q_{\beta}=\mathrm{10^{-4}}I_{2}$.
Targets appear with $\overline{x}_{a}=\left[450\,(\mathrm{km}),-0.1\,(\mathrm{km}/\mathrm{s}),6500\,(\mathrm{km}),-1\,(\mathrm{km}/\mathrm{s})\right]^{T}$
and $P_{a}=$
\[
\mathrm{diag}\left(\left[100^{2}(\mathrm{km}^{2}),3^{2}(\mathrm{\mathrm{km}^{2}}/\mathrm{s}^{2}),50^{2}(\mathrm{km}^{2}),1(\mathrm{km}^{2}/\mathrm{s}^{2})\right]\right).
\]
The parameters that model the appearance and disappearance times of
targets are $\lambda=0.3\,\mathrm{s}^{-1}$ and $\mu=0.01\,\mathrm{s}^{-1}$. 

As in the previous scenario, the sensor takes 100 measurements with
the same time interval $\Delta t_{k}$ between measurements. The sensor
measures positions with $p^{D}=0.9$ and matrices (\ref{eq:measurement_model_scenario1})
with $\sigma_{r}^{2}=10^{-2}\mathrm{km^{2}}$. The clutter intensity
is $\lambda^{C}\left(z\right)=10u_{A}\left(z\right)$ with area $A=\left[0,900\right]\times\left[0,6600\right]\,\left(\mathrm{km^{2}}\right)$.

\begin{figure}
\begin{centering}
\includegraphics[scale=0.6]{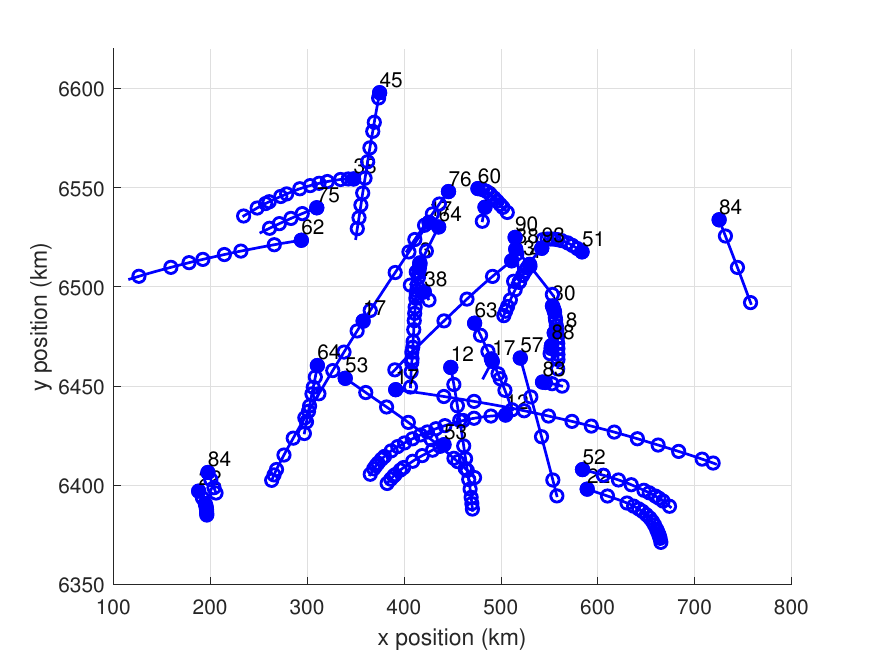}
\par\end{centering}
\caption{\label{fig:Ground-truth-trajectories-nonlinear}Ground truth set of
trajectories}
\end{figure}

\begin{figure}
\begin{centering}
\includegraphics[scale=0.3]{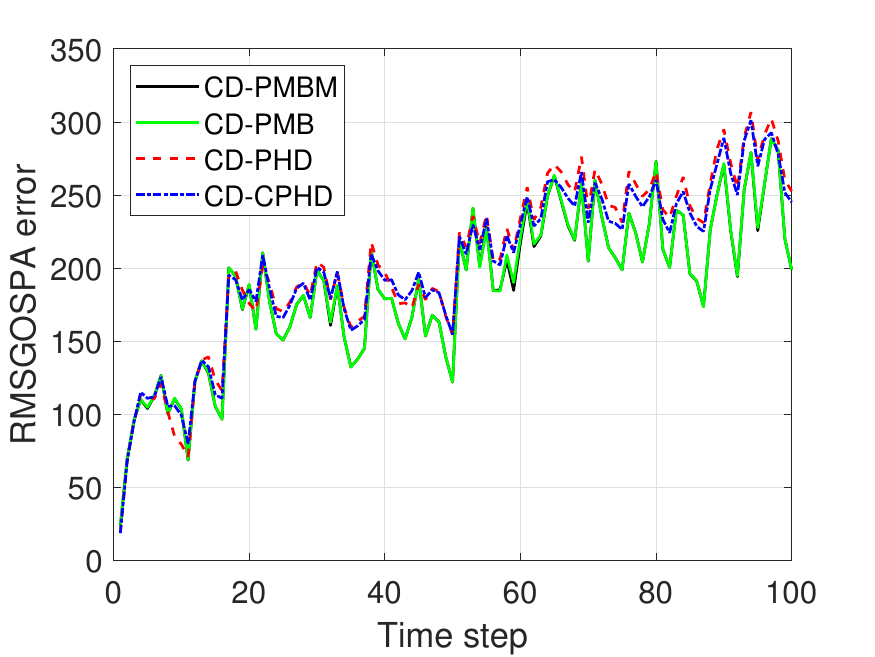}\includegraphics[scale=0.3]{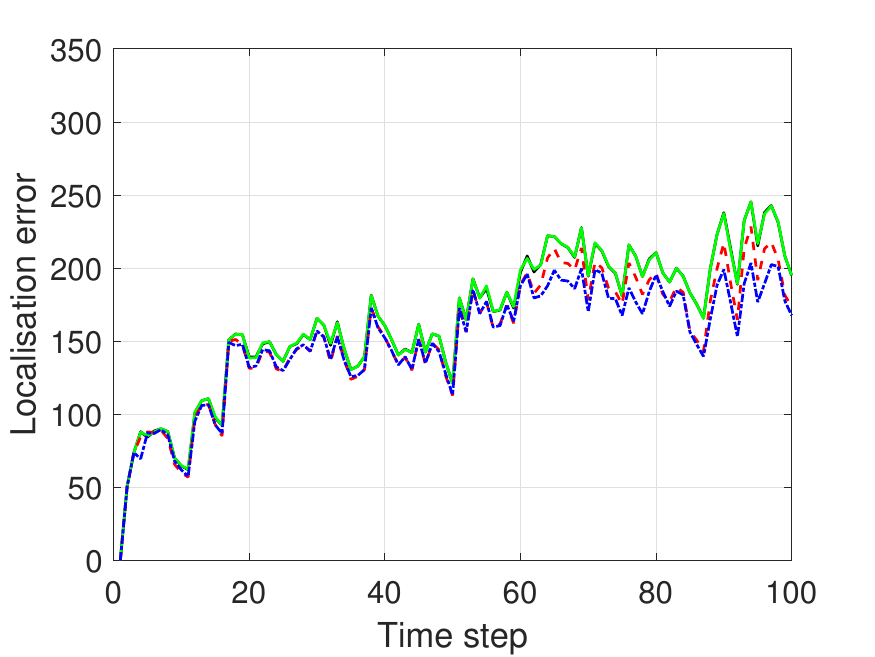}
\par\end{centering}
\begin{centering}
\includegraphics[scale=0.3]{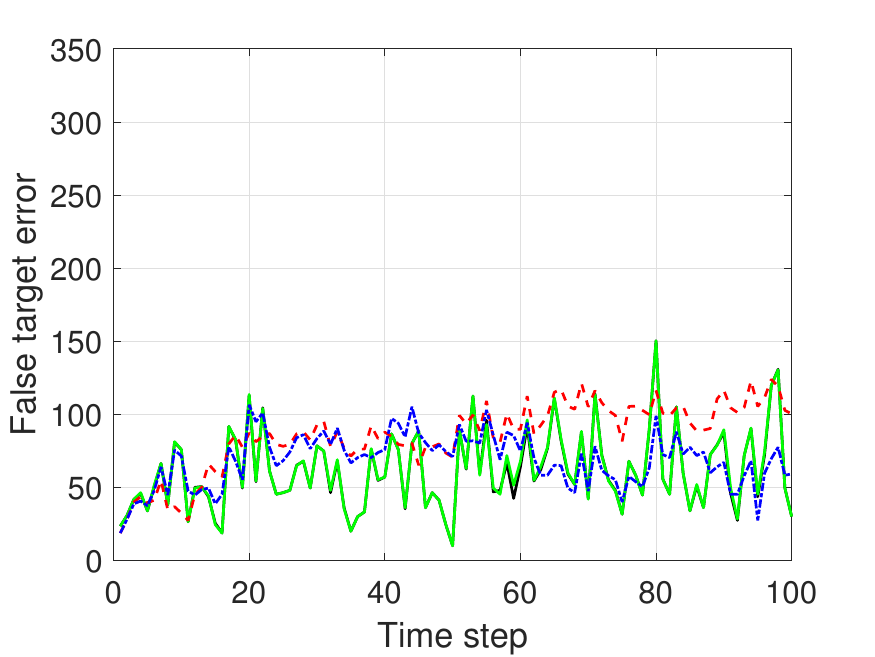}\includegraphics[scale=0.3]{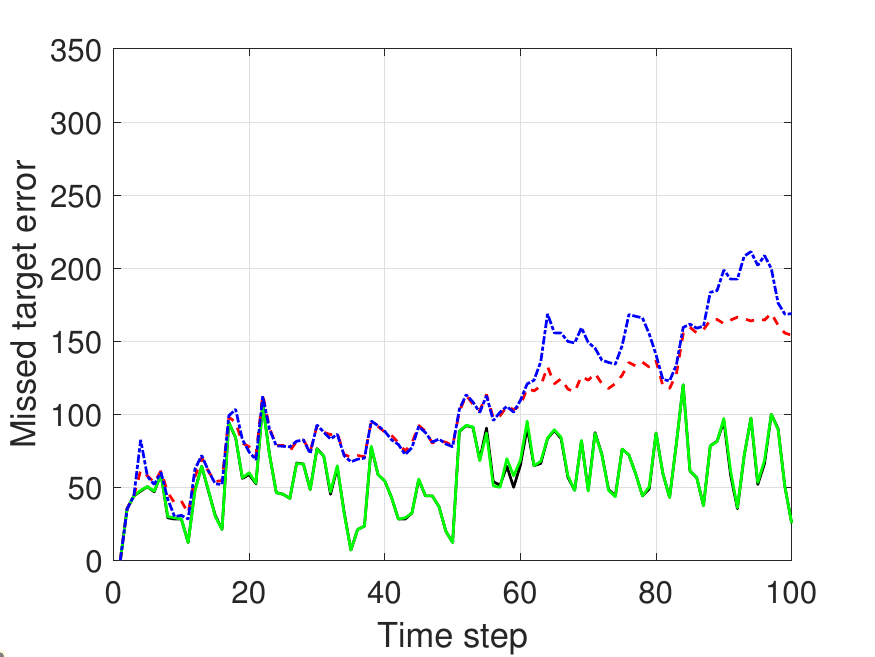}
\par\end{centering}
\caption{\label{fig:RMS-GOSPA-nonlinear}RMS GOSPA positional errors ($\mathrm{m)}$
and their decomposition against time.}

\end{figure}

We sample the ground truth set of trajectories at the sensor sampling
times using an Euler-Maruyama method to simulate the trajectories
from the ODEs with a step-size of 0.01 s \cite[Alg. 8.1]{Sarkka_book19}.
The resulting set of trajectories is shown in Figure \ref{fig:Ground-truth-trajectories-nonlinear}.
The parameters of the GOSPA metric in this scenario are $\alpha=2$,
$c=0.1\,\mathrm{km}$ and $p=2$. The RMS-GOSPA error and its decomposition
at each time step are shown in Figure \ref{fig:RMS-GOSPA-nonlinear}.
CD-PMBM and CD-PMB filters have the best performance. The worse performance
of CD-PHD and CD-CPHD filters is mainly due to an increase in the
number of missed targets, though these filters also usually report
a higher number of false targets.

As in the linear case, the PMBM, PMB, PHD and CPHD filters have very
similar errors to the corresponding continuous-discrete filters, so
the results are not shown in Figure \ref{fig:RMS-GOSPA-nonlinear}.
The computational times to run one Monte Carlo iteration of the filters
on an Intel core i5 laptop are: 2.6 s (CD-PMBM), 2.4 s (CD-PMB), 3.3
s (CD-PHD), 3.3 s (CD-CPHD), 9424.1 s (PMBM), 8800.4 s (PMB), 307.3
s (PHD), 319.6 s (CPHD). The continuous-discrete filters are remarkably
faster than the discrete filters. For instance, the CD-PMBM filter
is over 3500 times faster than the PMBM filter.

The results for the different single-target birth density discretisations,
explained in Section \ref{subsec:Linear-SDE-simulation}, are quire
similar in this scenario. As in the previous scenario, we analyse
the effect for higher mean appearing velocities $\overline{v}_{a}=[10,-10]^{T}\,(\mathrm{km}/\mathrm{s})$
and $\overline{v}_{a}=[15,-15]^{T}\,(\mathrm{km}/\mathrm{s})$, and
also for an informative birth density $P_{a}=I_{4}$, and also a constant
$\Delta t_{k}=1\mathrm{s}$.

Table \ref{tab:RMSGOSPA-non-linear} shows the RMS-GOSPA error across
all time steps for the different single-target birth density discretisations
in the four considered scenarios. As before, the birth discretisation
of Prop. 8 outperforms the other two discretisations. The improvement
is more significant in the informative birth cases.

\begin{table*}
\caption{\label{tab:RMSGOSPA-non-linear}RMS-GOSPA error (m) across all time
steps for different single-target birth discretisations in the non-linear
SDE example}

\centering{}%
\begin{tabular}{c|ccc|ccc|ccc|ccc}
\hline 
 &
\multicolumn{3}{c|}{$\overline{v}_{a}=[10,-10]^{T}$} &
\multicolumn{3}{c|}{$\overline{v}_{a}=[15,-15]^{T}$} &
\multicolumn{3}{c|}{$\overline{v}_{a}=[15,-15]^{T}$,$P_{a}=I_{4}$} &
\multicolumn{3}{c}{$\overline{v}_{a}=[15,-15]^{T}$,$P_{a}=I_{4}$, $\Delta t_{k}=1\mathrm{s}$}\tabularnewline
\hline 
Birth &
Prop. 8 &
CSBD1 &
CSBD2 &
Prop. 8 &
CSBD1 &
CSBD2 &
Prop. 8 &
CSBD1 &
CSBD2 &
Prop. 8 &
CSBD1 &
CSBD2\tabularnewline
\hline 
CD-PMBM &
194.40 &
196.11 &
196.14 &
199.69 &
199.89 &
200.15 &
\uline{184.99} &
219.84 &
230.59 &
\uline{186.32} &
208.24 &
214.85\tabularnewline
CD-PMB &
\uline{193.71} &
195.56 &
195.62 &
\uline{199.68} &
199.86 &
200.11 &
185.17 &
220.82 &
231.28 &
186.82 &
208.99 &
214.88\tabularnewline
CD-PHD &
218.82 &
218.96 &
219.17 &
221.61 &
221.54 &
221.90 &
214.98 &
232.66 &
235.51 &
215.32 &
229.89 &
229.70\tabularnewline
CD-CPHD &
219.31 &
219.46 &
219.28 &
222.41 &
222.35 &
222.43 &
215.16 &
230.75 &
234.77 &
214.35 &
228.93 &
228.08\tabularnewline
\hline 
\end{tabular}
\end{table*}

\section{Conclusions\label{sec:Conclusions}}

This paper has derived the Gaussian implementation of the continuous-discrete
Poisson multi-Bernoulli mixture filter, which computes the filtering
density when target appearance, dynamics and disappearance are in
continuous time, and measurements are taken at discrete time steps.
We have derived the single-target transition density and the best
Gaussian fit, obtained via moment matching, to the single-target birth
distribution when dynamics are modelled by linear, time-invariant
SDEs. We have also derived the steady-state birth density, and the
corresponding filter simplification. 

The results have been extended to non-linear, time-invariant SDEs.
These yield the continuous-discrete versions of the PMBM, PMB, PHD
and CPHD filters. For non-uniform sampling intervals, the developed
continuous-discrete filters show similar estimation error with a remarkable
decrease in computational time compared to analogous discrete filters
with uniform sampling time.

There are many lines of possible future research based on this work,
for example, the derivation of continuous-discrete PMBM filters using
the exponential family of distributions \cite{Brigo99} or Gaussian
mixtures \cite{Lambert22b}, and the use of variational techniques
for continuous-discrete multi-target filtering.

\bibliographystyle{IEEEtran}
\bibliography{14C__Trabajo_laptop_Referencias_Referencias}

\begin{thebibliography}{10}
\providecommand{\url}[1]{#1}
\csname url@samestyle\endcsname
\providecommand{\newblock}{\relax}
\providecommand{\bibinfo}[2]{#2}
\providecommand{\BIBentrySTDinterwordspacing}{\spaceskip=0pt\relax}
\providecommand{\BIBentryALTinterwordstretchfactor}{4}
\providecommand{\BIBentryALTinterwordspacing}{\spaceskip=\fontdimen2\font plus
\BIBentryALTinterwordstretchfactor\fontdimen3\font minus
  \fontdimen4\font\relax}
\providecommand{\BIBforeignlanguage}[2]{{%
\expandafter\ifx\csname l@#1\endcsname\relax
\typeout{** WARNING: IEEEtran.bst: No hyphenation pattern has been}%
\typeout{** loaded for the language `#1'. Using the pattern for}%
\typeout{** the default language instead.}%
\else
\language=\csname l@#1\endcsname
\fi
#2}}
\providecommand{\BIBdecl}{\relax}
\BIBdecl

\bibitem{Challa_book11}
S.~Challa, M.~R. Morelande, D.~Musicki, and R.~J. Evans, \emph{Fundamentals of
  Object Tracking}.\hskip 1em plus 0.5em minus 0.4em\relax Cambridge University
  Press, 2011.

\bibitem{Meyer18}
F.~{Meyer}, T.~{Kropfreiter}, J.~L. {Williams}, R.~{Lau}, F.~{Hlawatsch},
  P.~{Braca}, and M.~Z. {Win}, ``Message passing algorithms for scalable
  multitarget tracking,'' \emph{Proceedings of the IEEE}, vol. 106, no.~2, pp.
  221--259, Feb. 2018.

\bibitem{Houssineau18}
J.~Houssineau and D.~E. Clark, ``Multitarget filtering with linearized
  complexity,'' \emph{IEEE Transactions on Signal Processing}, vol.~66, no.~18,
  pp. 4957--4970, Sept. 2018.

\bibitem{Delande19}
E.~Delande, J.~Houssineau, J.~Franco, C.~Frueh, D.~Clark, and M.~Jah, ``A new
  multi-target tracking algorithm for a large number of orbiting objects,''
  \emph{Advances in Space Research}, vol.~64, pp. 645--667, 2019.

\bibitem{Pang21}
S.~Pang and H.~Radha, ``Multi-object tracking using {P}oisson multi-{B}ernoulli
  mixture filtering for autonomous vehicles,'' in \emph{IEEE International
  Conference on Acoustics, Speech and Signal Processing}, 2021, pp. 7963--7967.

\bibitem{Kufoalor20}
D.~K.~M. Kufoalor, T.~A. Johansen, E.~F. Brekke, A.~Heps{\o}, and K.~Trnka,
  ``Autonomous maritime collision avoidance: Field verification of autonomous
  surface vehicle behavior in challenging scenarios,'' \emph{Journal of Field
  Robotics}, vol.~37, no.~3, 2020.

\bibitem{Lambert22}
M.~Lambert, S.~r. Bonnabel, and F.~Bach, ``The continuous-discrete variational
  {K}alman filter {(CD-VKF)},'' in \emph{IEEE 61st Conference on Decision and
  Control}, 2022, pp. 6632--6639.

\bibitem{Sarkka_book19}
S.~S\"arkk\"a and A.~Solin, \emph{Applied Stochastic Differential
  Equations}.\hskip 1em plus 0.5em minus 0.4em\relax Cambridge University
  Press, 2019.

\bibitem{Hu10}
Y.~Hu, Z.~Duan, and D.~Zhou, ``Estimation fusion with general asynchronous
  multi-rate sensors,'' \emph{IEEE Transactions on Aerospace and Electronic
  Systems}, vol.~46, no.~4, pp. 2090--2102, 2010.

\bibitem{Eng_thesis07}
F.~Eng, ``Non-uniform sampling in statistical signal processing,'' Ph.D.
  dissertation, Link{\"o}ping University, 2007.

\bibitem{Zhang06}
X.~Zhang, P.~Willett, and Y.~Bar-Shalom, ``Uniform versus nonuniform sampling
  when tracking in clutter,'' \emph{IEEE Transactions on Aerospace and
  Electronic Systems}, vol.~42, no.~2, pp. 388--400, 2006.

\bibitem{Ge20c}
X.~Ge, Q.-L. Han, X.-M. Zhang, L.~Ding, and F.~Yang, ``Distributed
  event-triggered estimation over sensor networks: A survey,'' \emph{IEEE
  Transactions on Cybernetics}, vol.~50, no.~3, pp. 1306--1320, 2020.

\bibitem{Uney19}
M.~{ ney}, L.~M. {Millefiori}, and P.~{Braca}, ``Data driven vessel trajectory
  forecasting using stochastic generative models,'' in \emph{IEEE International
  Conference on Acoustics, Speech and Signal Processing (ICASSP)}, 2019, pp.
  8459--8463.

\bibitem{Oksendal_book03}
B.~{\O}ksendal, \emph{Stochastic Differential Equations: An Introduction with
  Applications}.\hskip 1em plus 0.5em minus 0.4em\relax Springer-Verlag, 2003.

\bibitem{Li03}
X.~Li and V.~Jilkov, ``Survey of maneuvering target tracking. {P}art {I}:
  Dynamic models,'' \emph{IEEE Transactions on Aerospace and Electronic
  Systems}, vol.~39, no.~4, pp. 1333--1364, Oct. 2003.

\bibitem{Millefiori16}
L.~M. {Millefiori}, P.~{Braca}, and P.~{Willett}, ``Consistent estimation of
  randomly sampled {O}rnstein-{U}hlenbeck process long-run mean for long-term
  target state prediction,'' \emph{IEEE Signal Processing Letters}, vol.~23,
  no.~11, pp. 1562--1566, Nov. 2016.

\bibitem{Millefiori16b}
L.~M. {Millefiori}, P.~{Braca}, K.~{Bryan}, and P.~{Willett}, ``Modeling vessel
  kinematics using a stochastic mean-reverting process for long-term
  prediction,'' \emph{IEEE Transactions on Aerospace and Electronic Systems},
  vol.~52, no.~5, pp. 2313--2330, Oct. 2016.

\bibitem{Coraluppi12}
S.~Coraluppi and C.~Carthel, ``Stability and stationarity in target kinematic
  modeling,'' in \emph{IEEE Aerospace Conference}, 2012, pp. 1--8.

\bibitem{Morelande05b}
M.~Morelande and N.~Gordon, ``Target tracking through a coordinated turn,'' in
  \emph{Proceedings of IEEE International Conference on Acoustics, Speech, and
  Signal Processing}, vol.~4, 2005, pp. 21--24.

\bibitem{Tronarp19}
F.~Tronarp and S.~S{ }rkk{ }, ``Iterative statistical linear regression for
  {G}aussian smoothing in continuous-time non-linear stochastic dynamic
  systems,'' \emph{Signal Processing}, vol. 159, pp. 1--12, June 2019.

\bibitem{Axelsson15}
P.~{Axelsson} and F.~{Gustafsson}, ``Discrete-time solutions to the
  continuous-time differential {L}yapunov equation with applications to
  {K}alman filtering,'' \emph{IEEE Transactions on Automatic Control}, vol.~60,
  no.~3, pp. 632--643, March 2015.

\bibitem{Van_Loan78}
C.~Van~Loan, ``Computing integrals involving the matrix exponential,''
  \emph{IEEE Transactions on Automatic Control}, vol.~23, no.~3, pp. 395--404,
  1978.

\bibitem{Sarkka12b}
S.~S\"arkk\"a and A.~Solin, ``On continuous-discrete cubature {K}alman
  filtering,'' in \emph{16th IFAC Symposium on System Identification}, 2012,
  pp. 1221--1226.

\bibitem{Arasaratnam10}
I.~Arasaratnam, S.~Haykin, and T.~Hurd, ``Cubature {K}alman filtering for
  continuous-discrete systems: Theory and simulations,'' \emph{IEEE
  Transactions on Signal Processing}, vol.~58, no.~10, pp. 4977--4993, Oct.
  2010.

\bibitem{Angel20}
A.~F. Garc\'{i}a-Fern\'{a}ndez and S.~Maskell, ``Continuous-discrete multiple
  target filtering: {PMBM}, {PHD} and {CPHD} filter implementations,''
  \emph{IEEE Transactions on Signal Processing}, vol.~68, pp. 1300--1314, 2020.

\bibitem{Angel21_d}
A.~F. Garc\'{i}a-Fern\'{a}ndez and W.~Yi, ``Continuous-discrete multiple target
  tracking with out-of-sequence measurements,'' \emph{IEEE Transactions on
  Signal Processing}, vol.~69, pp. 4699--4709, 2021.

\bibitem{Coraluppi14}
S.~Coraluppi and C.~A. Carthel, ``If a tree falls in the woods, it does make a
  sound: multiple-hypothesis tracking with undetected target births,''
  \emph{IEEE Transactions on Aerospace and Electronic Systems}, vol.~50, no.~3,
  pp. 2379--2388, July 2014.

\bibitem{Kleinrock_book76}
L.~Kleinrock, \emph{Queueing Systems}.\hskip 1em plus 0.5em minus 0.4em\relax
  John Wiley \& Sons, 1976.

\bibitem{Williams15b}
J.~L. Williams, ``Marginal multi-{B}ernoulli filters: {RFS} derivation of
  {MHT}, {JIPDA} and association-based {MeMBer},'' \emph{IEEE Transactions on
  Aerospace and Electronic Systems}, vol.~51, no.~3, pp. 1664--1687, July 2015.

\bibitem{Angel18_b}
A.~F. Garc\'{i}a-Fern\'{a}ndez, J.~L. Williams, K.~Granstr{\"o}m, and
  L.~Svensson, ``Poisson multi-{B}ernoulli mixture filter: direct derivation
  and implementation,'' \emph{IEEE Transactions on Aerospace and Electronic
  Systems}, vol.~54, no.~4, pp. 1883--1901, Aug. 2018.

\bibitem{Brekke18}
E.~Brekke and M.~Chitre, ``Relationship between finite set statistics and the
  multiple hypothesis tracker,'' \emph{IEEE Transactions on Aerospace and
  Electronic Systems}, vol.~54, no.~4, pp. 1902--1917, Aug. 2018.

\bibitem{Bostrom-Rost21}
P.~Bostr{\"o}m-Rost, D.~{Axehill}, and G.~{Hendeby}, ``Sensor management for
  search and track using the {P}oisson multi-{B}ernoulli mixture filter,''
  \emph{IEEE Transactions on Aerospace and Electronic Systems}, vol.~57, no.~5,
  pp. 2771--2783, 2021.

\bibitem{Horwood11}
J.~T. Horwood, N.~D. Aragon, and A.~B. Poore, ``Gaussian sum filters for space
  surveillance: Theory and simulations,'' \emph{Journal of Guidance, Control
  and Dynamics}, vol.~34, pp. 1839--1851, 2011.

\bibitem{Crouse15}
D.~Crouse, ``Basic tracking using nonlinear continuous-time dynamic models
  [tutorial],'' \emph{IEEE Aerospace and Electronic Systems Magazine}, vol.~30,
  no.~2, pp. 4--41, 2015.

\bibitem{Mahler_book14}
R.~P.~S. Mahler, \emph{Advances in Statistical Multisource-Multitarget
  Information Fusion}.\hskip 1em plus 0.5em minus 0.4em\relax Artech House,
  2014.

\bibitem{Angel20_e}
A.~F. Garc\'{i}a-Fern\'{a}ndez, L.~Svensson, J.~L. Williams, Y.~Xia, and
  K.~Granstr{\"o}m, ``Trajectory {P}oisson multi-{B}ernoulli filters,''
  \emph{IEEE Transactions on Signal Processing}, vol.~68, pp. 4933--4945, 2020.

\bibitem{Williams15}
J.~L. Williams, ``An efficient, variational approximation of the best fitting
  multi-{B}ernoulli filter,'' \emph{IEEE Transactions on Signal Processing},
  vol.~63, no.~1, pp. 258--273, Jan. 2015.

\bibitem{Angel15_d}
A.~F. Garc\'{i}a-Fern\'{a}ndez and B.-N. Vo, ``Derivation of the {PHD} and
  {CPHD} filters based on direct {Kullback-Leibler} divergence minimization,''
  \emph{IEEE Transactions on Signal Processing}, vol.~63, no.~21, pp.
  5812--5820, Nov. 2015.

\bibitem{Carbonell08}
F.~Carbonell, J.~C.~J. menez, and L.~M. Pedroso, ``Computing multiple integrals
  involving matrix exponentials,'' \emph{Journal of Computational and Applied
  Mathematics}, vol. 213, pp. 300--305, 2008.

\bibitem{Kulkarni_book16}
V.~G. Kulkarni, \emph{Modeling and analysis of stochastic systems}.\hskip 1em
  plus 0.5em minus 0.4em\relax Chapman \& Hall/CRC, 2016.

\bibitem{Hershey07}
J.~Hershey and P.~Olsen, ``Approximating the {K}ullback {L}eibler divergence
  between {G}aussian mixture models,'' in \emph{IEEE International Conference
  on Acoustics, Speech and Signal Processing}, vol.~4, April 2007, pp.
  317--320.

\bibitem{Golub_book96}
G.~H. Golub and C.~F. Van~Loan, \emph{Matrix computations}.\hskip 1em plus
  0.5em minus 0.4em\relax The Johns Hopkins University Press, 1996.

\bibitem{Khalil_book02}
H.~K. Khalil, \emph{Nonlinear systems}.\hskip 1em plus 0.5em minus 0.4em\relax
  Prentice Hall, 2002.

\bibitem{Vo06}
B.-N. Vo and W.-K. Ma, ``The {G}aussian mixture probability hypothesis density
  filter,'' \emph{IEEE Transactions on Signal Processing}, vol.~54, no.~11, pp.
  4091--4104, Nov. 2006.

\bibitem{Vo07}
B.-T. Vo, B.-N. Vo, and A.~Cantoni, ``Analytic implementations of the
  cardinalized probability hypothesis density filter,'' \emph{IEEE Transactions
  on Signal Processing}, vol.~55, no.~7, pp. 3553--3567, July 2007.

\bibitem{Granstrom20}
K.~Granstr{\"o}m, M.~Fatemi, and L.~Svensson, ``Poisson multi-{B}ernoulli
  mixture conjugate prior for multiple extended target filtering,'' \emph{IEEE
  Transactions on Aerospace and Electronic Systems}, vol.~56, no.~1, pp.
  208--225, Feb. 2020.

\bibitem{Rahmathullah17}
A.~S. Rahmathullah, A.~F.~G. a~Fern~ndez, and L.~Svensson, ``Generalized
  optimal sub-pattern assignment metric,'' in \emph{20th International
  Conference on Information Fusion}, 2017, pp. 1--8.

\bibitem{Julier04}
S.~J. Julier and J.~K. Uhlmann, ``Unscented filtering and nonlinear
  estimation,'' \emph{Proceedings of the IEEE}, vol.~92, no.~3, pp. 401--422,
  Mar. 2004.

\bibitem{Brigo99}
D.~Brigo, B.~Hanzon, and F.~{Le Gland}, ``Approximate nonlinear filtering by
  projection on exponential manifolds of densities,'' \emph{Bernoulli}, vol.~5,
  no.~3, pp. 495--534, 1999.

\bibitem{Lambert22b}
M.~Lambert, S.~Chewi, F.~Bach, S.~re~Bonnabel, and P.~Rigollet, ``Variational
  inference via {W}asserstein gradient flows,'' in \emph{36th Conference on
  Neural Information Processing Systems}, 2022, pp. 1--14.

\bibitem{Hall_book15}
B.~C. Hall, \emph{Lie groups, Lie algebras, and representations: An elementary
  introduction}.\hskip 1em plus 0.5em minus 0.4em\relax Springer, 2015.

\bibitem{Ross_book10}
S.~Ross, \emph{A First Course in Probability}, 8th~ed.\hskip 1em plus 0.5em
  minus 0.4em\relax Prentice-Hall, 2010.

\bibitem{Apostol_book67}
T.~M. Apostol, \emph{Calculus. Volume I.}\hskip 1em plus 0.5em minus
  0.4em\relax John Wiley \& Sons, 1967.

\end{thebibliography}

\cleardoublepage{}

{\LARGE Supplemental material: ``Gaussian multi-target filtering with
target dynamics driven by a stochastic differential equation''}{\LARGE\par}

\appendices{}

\section{\label{sec:Appendix_A}}

In this appendix we prove (\ref{eq:integral_type3}). For the given
$H_{c}^{3}$ in (\ref{eq:H_c3}),  the value of $H_{1}(t)$ in (\ref{eq:expm_int_type2})
is given by \cite[Sec. II]{Van_Loan78}
\begin{align}
H_{1}(t) & =\int_{0}^{t}\int_{0}^{s}\exp\left(-B\left(t-s\right)\right)Q_{c}\exp\left(Ar\right)drds\nonumber \\
 & =\exp\left(-Bt\right)\int_{0}^{t}\exp\left(Bs\right)Q_{c}\left(\int_{0}^{s}\exp\left(Ar\right)dr\right)ds.\label{eq:H_1_append}
\end{align}
Therefore, by using the property that, for a square matrix $X$ \cite{Hall_book15},
\begin{align}
\exp\left(-X\right) & =\left(\exp\left(X\right)\right)^{-1},
\end{align}
and multiplying by the inverse of $\exp\left(-Bt\right)$ on the left
for both sides of the equation in (\ref{eq:H_1_append}), we complete
the proof of (\ref{eq:integral_type3}).

\section{\label{sec:Appendix_B}}

In this appendix, we prove Proposition \ref{prop:mean_cov_birth}.
We prove the mean in Section \ref{subsec:Proof-Mean} and the covariance
matrix in Section \ref{subsec:Proof_Covariance-matrix}. 

It should be noted that (\ref{eq:mean_birth_prev}) and (\ref{eq:cov_birth})
are obtained by the law of total expectation and the law of total
covariance, respectively \cite{Ross_book10}. In the proofs, we will
use this result. Given two $n_{x}\text{\ensuremath{\times n_{x}}}$
matrices $X$ and $Y$ such that $XY=YX$, then \cite{Hall_book15}
\begin{align}
\exp\left(X\right)\exp\left(Y\right) & =\exp\left(X+Y\right).\label{eq:product_matrix_exponentials}
\end{align}

\subsection{Mean\label{subsec:Proof-Mean}}

Plugging (\ref{eq:density_time_lag}) and (\ref{eq:E_x_t}) into (\ref{eq:mean_birth_prev})
yields
\begin{align}
\overline{x}_{b,k} & =\overline{x}_{b,k,1}+\overline{x}_{b,k,2},
\end{align}
where
\begin{align}
\overline{x}_{b,k,1} & =\frac{\mu}{1-e^{-\mu\Delta t_{k}}}\int_{0}^{\Delta t_{k}}\exp\left(At\right)\overline{x}_{a}e^{-\mu t}dt,\label{eq:L_1_mean_append}\\
\overline{x}_{b,k,2} & =\frac{\mu}{1-e^{-\mu\Delta t_{k}}}\int_{0}^{\Delta t_{k}}\left[\int_{0}^{t}\exp\left(A\tau\right)d\tau\right]ue^{-\mu t}dt.\label{eq:L_2_mean_append}
\end{align}
We proceed to calculate (\ref{eq:L_1_mean_append}) and (\ref{eq:L_2_mean_append}).

\subsubsection{Calculation of (\ref{eq:L_1_mean_append})\label{subsec:Calculation-of-L1}}

Operating on (\ref{eq:L_1_mean_append}), we obtain
\begin{align}
\overline{x}_{b,k,1} & =\frac{\mu}{1-e^{-\mu\Delta t_{k}}}\left[\int_{0}^{\Delta t_{k}}\exp\left(At\right)e^{-\mu t}Idt\right]\overline{x}_{a}\nonumber \\
 & =\frac{\mu}{1-e^{-\mu\Delta t_{k}}}\left[\int_{0}^{\Delta t_{k}}\exp\left(At\right)e^{-\mu It}dt\right]\overline{x}_{a}.
\end{align}
We know that $At\times\mu I_{n_{x}}t=\mu I_{n_{x}}t\times At$, then
we can apply (\ref{eq:product_matrix_exponentials}) to obtain
\begin{align}
\overline{x}_{b,k,1} & =\frac{\mu}{1-e^{-\mu\Delta t_{k}}}\left[\int_{0}^{\Delta t_{k}}\exp\left(\left(A-\mu I\right)t\right)dt\right]\overline{x}_{a}.
\end{align}
Making use of (\ref{eq:Integral_type1}) yields
\begin{align}
\overline{x}_{b,k,1} & =\frac{\mu}{1-e^{-\mu\Delta t_{k}}}\left[\begin{array}{cc}
I & 0_{n_{x},1}\end{array}\right]\exp\left(\overline{A}_{b}\Delta t_{k}\right)\left[\begin{array}{c}
0_{n_{x},1}\\
1
\end{array}\right],
\end{align}
where $\overline{A}_{b}$ is given by (\ref{eq:mean_birth_Ab}). 

\subsubsection{Calculation of (\ref{eq:L_2_mean_append})}

We first calculate the inner integral in (\ref{eq:L_2_mean_append}).
Using (\ref{eq:Integral_type1}), we obtain
\begin{align}
\int_{0}^{t}\exp\left(A\tau\right)d\tau u & =\left[\begin{array}{cc}
I & 0_{n_{x},1}\end{array}\right]\exp\left(\overline{A}_{b,u,1}t\right)\left[\begin{array}{c}
0_{n_{x},1}\\
1
\end{array}\right],
\end{align}
where
\begin{align}
\overline{A}_{b,u,1} & =\left[\begin{array}{cc}
A & u\\
0_{1,n_{x}} & 0
\end{array}\right].
\end{align}
Substituting this expression into (\ref{eq:L_2_mean_append}) yields
\begin{align}
\overline{x}_{b,k,2} & =\frac{\mu}{1-e^{-\mu\Delta t_{k}}}\left[\begin{array}{cc}
I & 0_{n_{x},1}\end{array}\right]\nonumber \\
 & \quad\times\int_{0}^{\Delta t_{k}}\exp\left(\overline{A}_{b,u,1}t\right)e^{-\mu t}dt\left[\begin{array}{c}
0_{n_{x},1}\\
1
\end{array}\right].
\end{align}

We calculate the following integral using the approach in Section
\ref{subsec:Calculation-of-L1}
\begin{align}
 & \int_{0}^{\Delta t_{k}}\exp\left(\overline{A}_{b,u,1}t\right)e^{-\mu t}dt\left[\begin{array}{c}
0_{n_{x},1}\\
1
\end{array}\right]\nonumber \\
 & =\int_{0}^{\Delta t_{k}}\exp\left(\left(\overline{A}_{b,u,1}-\mu I_{n_{x}+1}\right)t\right)dt\left[\begin{array}{c}
0_{n_{x},1}\\
1
\end{array}\right]\nonumber \\
 & =\int_{0}^{\Delta t_{k}}\exp\left(\left[\begin{array}{cc}
A-\mu I & u\\
0_{1,n_{x}} & -\mu
\end{array}\right]t\right)dt\left[\begin{array}{c}
0_{n_{x},1}\\
1
\end{array}\right],
\end{align}
where $I_{n_{x}+1}$ is an identity matrix of size $n_{x}+1$.

Making use of (\ref{eq:Integral_type1}), we obtain
\begin{align}
\overline{x}_{b,k,2} & =\frac{\mu}{1-e^{-\mu\Delta t_{k}}}\left[\begin{array}{cc}
I & 0_{n_{x},1}\end{array}\right]\left[\begin{array}{cc}
I_{n_{x}+1} & 0_{n_{x}+1,1}\end{array}\right]\nonumber \\
 & \quad\times\exp\left(\overline{A}_{b,u,2}t\right)\left[\begin{array}{c}
0_{n_{x}+1,1}\\
1
\end{array}\right]\nonumber \\
 & =\frac{\mu}{1-e^{-\mu\Delta t_{k}}}\left[\begin{array}{cc}
I & 0_{n_{x},2}\end{array}\right]\exp\left(\overline{A}_{b,u}t\right)\left[\begin{array}{c}
0_{n_{x}+1,1}\\
1
\end{array}\right],
\end{align}
where $\overline{A}_{b,u}$ is given by (\ref{eq:mean_birth_Abu2}).
The calculated values of $\overline{x}_{b,k,1}$ and $\overline{x}_{b,k,2}$
finish the proof of (\ref{eq:mean_birth}). 

\subsection{Term $\mathrm{E}\left[\mathrm{C}\left[x_{k}\left|t\right.\right]\right]$\label{subsec:Proof_Covariance-matrix}}

We proceed to prove $\mathrm{E}\left[\mathrm{C}\left[x_{k}\left|t\right.\right]\right]$
in (\ref{eq:cov_birth}). Using (\ref{eq:density_time_lag}) and (\ref{eq:C_x_t}),
we write
\begin{align}
\mathrm{E}\left[\mathrm{C}\left[x_{k}\left|t\right.\right]\right] & =E_{1}+E_{2},\label{eq:E_C_x_append}
\end{align}
where
\begin{align}
E_{1} & =\mathrm{E}\left[\int_{0}^{t}\exp\left(A\tau\right)LQ_{\beta}L^{T}\exp\left(A^{T}\tau\right)d\tau\right],\nonumber \\
 & =\frac{\mu}{1-e^{-\mu\Delta t_{k}}}\int_{0}^{\Delta t_{k}}e^{-\mu t}\\
 & \quad\times\left[\int_{0}^{t}\exp\left(A\tau\right)LQ_{\beta}L^{T}\exp\left(A^{T}\tau\right)d\tau\right]dt,\label{eq:E_1_appendix}
\end{align}
and 
\begin{align}
E_{2} & =\mathrm{E}\left[\exp\left(At\right)P_{a}\exp\left(A^{T}t\right)\right],\nonumber \\
 & =\frac{\mu}{1-e^{-\mu\Delta t_{k}}}\int_{0}^{\Delta t_{k}}\exp\left(At\right)P_{a}\exp\left(A^{T}t\right)e^{-\mu t}dt.\label{eq:E_2_appendix}
\end{align}
We first calculate $E_{1}$ and then proceed to calculate (\ref{eq:E_C_x_append}). 

\subsubsection{Term $E_{1}$}

We calculate (\ref{eq:E_1_appendix}) by integrating by parts \cite{Apostol_book67}.
We define 
\begin{align*}
u(t) & =\frac{\mu}{1-e^{-\mu\Delta t_{k}}}\int_{0}^{t}\exp\left(A\tau\right)LQ_{\beta}L^{T}\exp\left(A^{T}\tau\right)d\tau,\\
v'(t)dt & =e^{-\mu t}dt,
\end{align*}
which implies
\begin{align*}
u'(t)dt & =\frac{\mu}{1-e^{-\mu\Delta t_{k}}}\exp\left(At\right)LQ_{\beta}L^{T}\exp\left(A^{T}t\right),\\
v(t) & =-\frac{e^{-\mu t}}{\mu}.
\end{align*}
Then,
\begin{align}
E_{1} & =\frac{1}{1-e^{-\mu\Delta t_{k}}}\left[-e^{-\mu\Delta t_{k}}\right.\nonumber \\
 & \quad\times\int_{0}^{\Delta t_{k}}\exp\left(A\tau\right)LQ_{\beta}L^{T}\exp\left(A^{T}\tau\right)d\tau\nonumber \\
 & \quad\left.+\int_{0}^{\Delta t_{k}}e^{-\mu\tau}\exp\left(A\tau\right)LQ_{\beta}L^{T}\exp\left(A^{T}\tau\right)d\tau\right].\label{eq:E_1_prev_append}
\end{align}

\subsubsection{Rest of the proof}

Substituting (\ref{eq:E_1_prev_append}) and (\ref{eq:E_2_appendix})
into (\ref{eq:E_C_x_append}), we obtain
\begin{align}
 & \mathrm{E}\left[\mathrm{C}\left[x_{k}\left|t\right.\right]\right]\nonumber \\
 & =-\frac{e^{-\mu\Delta t_{k}}}{1-e^{-\mu\Delta t_{k}}}\int_{0}^{\Delta t_{k}}\exp\left(A\tau\right)LQ_{\beta}L^{T}\exp\left(A^{T}\tau\right)d\tau\nonumber \\
 & +\frac{1}{1-e^{-\mu\Delta t_{k}}}\int_{0}^{\Delta t_{k}}e^{-\mu\tau}\exp\left(A\tau\right)C_{b}\exp\left(A^{T}\tau\right)d\tau,\label{eq:E_C_x_integral}
\end{align}
where $C_{b}$ is given by (\ref{eq:C_b}). The first integral in
(\ref{eq:E_C_x_integral}) can be calculated using (\ref{eq:Integral_type2}).
Following the procedure in Section \ref{subsec:Calculation-of-L1},
the second integral in (\ref{eq:E_C_x_integral}) can be written as
\begin{align}
 & \int_{0}^{\Delta t_{k}}e^{-\mu t}\exp\left(At\right)C_{b}\exp\left(A^{T}t\right)dt\nonumber \\
 & =\int_{0}^{\Delta t_{k}}\exp\left(\left(A-\mu/2I\right)t\right)C_{b}\exp\left(\left(A-\mu/2I\right)^{T}t\right)dt.\label{eq:E_C_x_integral2}
\end{align}
Integral (\ref{eq:E_C_x_integral2}) can be computed using (\ref{eq:Integral_type2}).
This proves the expression for $\mathrm{E}\left[\mathrm{C}\left[x_{k}\left|t\right.\right]\right]$
in Proposition \ref{prop:mean_cov_birth}.

\subsection{Term $\mathrm{C}\left[\mathrm{E}\left[x_{k}\left|t\right.\right]\right]$}

We expand $\mathrm{C}\left[\mathrm{E}\left[x_{k}\left|t\right.\right]\right]$
as
\begin{align}
\mathrm{C}\left[\mathrm{E}\left[x_{k}\left|t\right.\right]\right] & =\mathrm{E}\left[\mathrm{E}\left[x_{k}\left|t\right.\right]\mathrm{E}\left[x_{k}\left|t\right.\right]^{T}\right]-\overline{x}_{b,k}\overline{x}_{b,k}^{T}.
\end{align}
Substituting (\ref{eq:E_x_t}) into $\mathrm{E}\left[\mathrm{E}\left[x_{k}\left|t\right.\right]\mathrm{E}\left[x_{k}\left|t\right.\right]^{T}\right]$,
we obtain
\begin{align}
 & \mathrm{E}\left[\mathrm{E}\left[x_{k}\left|t\right.\right]\mathrm{E}\left[x_{k}\left|t\right.\right]^{T}\right]\nonumber \\
 & =\mathrm{E}\left[\left(\exp\left(At\right)\overline{x}_{a}+\int_{0}^{t}\exp\left(A\tau\right)d\tau u\right)\right.\nonumber \\
 & \left.\times\left(\overline{x}_{a}^{T}\exp\left(A^{T}t\right)+u^{T}\int_{0}^{t}\exp\left(A^{T}\tau\right)d\tau\right)\right]\nonumber \\
 & =\mathrm{E}\left[\exp\left(At\right)\overline{x}_{a}\overline{x}_{a}^{T}\exp\left(A^{T}t\right)\right]\nonumber \\
 & +\mathrm{E}\left[\exp\left(At\right)\overline{x}_{a}u^{T}\left(\int_{0}^{t}\exp\left(A^{T}\tau\right)d\tau\right)\right]\nonumber \\
 & +\mathrm{E}\left[\int_{0}^{t}\exp\left(A\tau\right)d\tau u\overline{x}_{a}^{T}\exp\left(A^{T}t\right)\right]\nonumber \\
 & +\mathrm{E}\left[\int_{0}^{t}\exp\left(A\tau\right)d\tau uu^{T}\int_{0}^{t}\exp\left(A^{T}\tau\right)d\tau\right].\label{eq:C_E_x_prov_append1}
\end{align}
We proceed to calculate the required integrals in (\ref{eq:C_E_x_prov_append1}).
The first type of integral in (\ref{eq:C_E_x_prov_append1}) is
\begin{align}
\Sigma_{xx} & =\mathrm{E}\left[\exp\left(At\right)\overline{x}_{a}\overline{x}_{a}^{T}\exp\left(A^{T}t\right)\right]\nonumber \\
 & =\frac{\mu}{1-e^{-\mu\Delta t_{k}}}\int_{0}^{\Delta t_{k}}\exp\left(At\right)\overline{x}_{a}\overline{x}_{a}^{T}\exp\left(A^{T}t\right)e^{-\mu t}dt\nonumber \\
 & =\frac{\mu}{1-e^{-\mu\Delta t_{k}}}\int_{0}^{\Delta t_{k}}\exp\left(\left(A-\mu/2I\right)t\right)\overline{x}_{a}\overline{x}_{a}^{T}\nonumber \\
 & \quad\times\exp\left(\left(A-\mu/2I\right)^{T}t\right)dt.
\end{align}
This is an integral of the form (\ref{eq:Integral_type2}) that we
can calculate. 

The second type of integral in (\ref{eq:C_E_x_prov_append1}) is
\begin{align}
\Sigma_{xu} & =\mathrm{E}\left[\exp\left(At\right)\overline{x}_{a}u^{T}\left(\int_{0}^{t}\exp\left(A^{T}\tau\right)d\tau\right)\right]\nonumber \\
 & =\frac{\mu}{1-e^{-\mu\Delta t_{k}}}\int_{0}^{\Delta t_{k}}\exp\left(At\right)\overline{x}_{a}u^{T}\nonumber \\
 & \quad\times\left(\int_{0}^{t}\exp\left(A^{T}\tau\right)d\tau\right)e^{-\mu t}dt\nonumber \\
 & =\frac{\mu}{1-e^{-\mu\Delta t_{k}}}\int_{0}^{\Delta t_{k}}\exp\left(\left(A-\mu I\right)t\right)\overline{x}_{a}u^{T}\nonumber \\
 & \quad\times\left(\int_{0}^{t}\exp\left(A^{T}\tau\right)d\tau\right)dt.
\end{align}
This is an integral of the form (\ref{eq:integral_type3}) that we
can calculate. 

The third type of integral in (\ref{eq:C_E_x_prov_append1}) is
\begin{align}
\Sigma_{uu} & =\mathrm{E}\left[\int_{0}^{t}\exp\left(A\tau\right)d\tau uu^{T}\int_{0}^{t}\exp\left(A^{T}\tau\right)d\tau\right]\nonumber \\
 & =\frac{\mu}{1-e^{-\mu\Delta t_{k}}}\int_{0}^{\Delta t_{k}}e^{-\mu t}\nonumber \\
 & \times\left[\int_{0}^{t}\exp\left(A\tau\right)d\tau uu^{T}\int_{0}^{t}\exp\left(A^{T}\tau\right)d\tau\right]dt.\label{eq:Sigma_uu_append}
\end{align}

\subsubsection{Integration by parts for $\Sigma_{uu}$}

We compute (\ref{eq:Sigma_uu_append}) by parts. We define
\begin{align}
u(t) & =\frac{\mu}{1-e^{-\mu\Delta t_{k}}}\int_{0}^{t}\exp\left(A\tau\right)d\tau uu^{T}\int_{0}^{t}\exp\left(A^{T}\tau\right)d\tau,\\
v'(t)dt & =e^{-\mu t}dt,
\end{align}
which implies
\begin{align}
u'(t)dt & =\frac{\mu}{1-e^{-\mu\Delta t_{k}}}\left[\exp\left(At\right)uu^{T}\left(\int_{0}^{t}\exp\left(A^{T}\tau\right)d\tau\right)\right.\nonumber \\
 & \left.+\left(\int_{0}^{t}\exp\left(A\tau\right)d\tau\right)uu^{T}\exp\left(A^{T}t\right)\right],\\
v(t) & =-\frac{e^{-\mu t}}{\mu}.
\end{align}

Then, we have
\begin{align}
 & \Sigma_{uu}\nonumber \\
 & =-\frac{\mu}{1-e^{-\mu\Delta t_{k}}}\nonumber \\
 & \times\left[\int_{0}^{t}\exp\left(A\tau\right)d\tau uu^{T}\int_{0}^{t}\exp\left(A^{T}\tau\right)d\tau\frac{e^{-\mu t}}{\mu}\right]_{0}^{\Delta t_{k}}\nonumber \\
 & +\frac{\mu}{1-e^{-\mu\Delta t_{k}}}\int_{0}^{\Delta t_{k}}\frac{e^{-\mu t}}{\mu}\nonumber \\
 & \times\left[\exp\left(At\right)uu^{T}\left(\int_{0}^{t}\exp\left(A^{T}\tau\right)d\tau\right)\right.\nonumber \\
 & =-\frac{\mu}{1-e^{-\mu\Delta t_{k}}}\left[\frac{e^{-\mu\Delta t_{k}}}{\mu}\int_{0}^{\Delta t_{k}}\exp\left(A\tau\right)d\tau uu^{T}\right.\nonumber \\
 & \left.\times\int_{0}^{\Delta t_{k}}\exp\left(A^{T}\tau\right)d\tau\right]\nonumber \\
 & +\frac{\mu}{1-e^{-\mu\Delta t_{k}}}\int_{0}^{\Delta t_{k}}\exp\left(At\right)uu^{T}\nonumber \\
 & \times\left(\int_{0}^{t}\exp\left(A^{T}\tau\right)d\tau\right)\frac{e^{-\mu t}}{\mu}dt\nonumber \\
 & +\frac{\mu}{1-e^{-\mu\Delta t_{k}}}\int_{0}^{\Delta t_{k}}\left(\int_{0}^{t}\exp\left(A\tau\right)d\tau\right)uu^{T}\nonumber \\
 & \times\exp\left(A^{T}t\right)\frac{e^{-\mu t}}{\mu}dt\nonumber \\
 & =-\frac{e^{-\mu\Delta t_{k}}}{1-e^{-\mu\Delta t_{k}}}\int_{0}^{\Delta t_{k}}\exp\left(A\tau\right)d\tau uu^{T}\nonumber \\
 & \times\int_{0}^{\Delta t_{k}}\exp\left(A^{T}\tau\right)d\tau\nonumber \\
 & +\frac{1}{1-e^{-\mu\Delta t_{k}}}\int_{0}^{\Delta t_{k}}\exp\left(At\right)uu^{T}\nonumber \\
 & \times\left(\int_{0}^{t}\exp\left(A^{T}\tau\right)d\tau\right)e^{-\mu t}dt\nonumber \\
 & +\frac{1}{1-e^{-\mu\Delta t_{k}}}\int_{0}^{\Delta t_{k}}\left(\int_{0}^{t}\exp\left(A\tau\right)d\tau\right)uu^{T}\nonumber \\
 & \times\exp\left(A^{T}t\right)e^{-\mu t}dt\nonumber \\
 & =-\frac{e^{-\mu\Delta t_{k}}}{1-e^{-\mu\Delta t_{k}}}\int_{0}^{\Delta t_{k}}\exp\left(A\tau\right)d\tau uu^{T}\nonumber \\
 & \times\int_{0}^{\Delta t_{k}}\exp\left(A^{T}\tau\right)d\tau\nonumber \\
 & +\frac{1}{1-e^{-\mu\Delta t_{k}}}\int_{0}^{\Delta t_{k}}\exp\left(\left(A-\mu I\right)t\right)uu^{T}\nonumber \\
 & \times\left(\int_{0}^{t}\exp\left(A^{T}\tau\right)d\tau\right)dt\nonumber \\
 & +\frac{1}{1-e^{-\mu\Delta t_{k}}}\int_{0}^{\Delta t_{k}}\left(\int_{0}^{t}\exp\left(A\tau\right)d\tau\right)uu^{T}\nonumber \\
 & \times\exp\left(\left(A-\mu I\right)^{T}t\right)dt.
\end{align}
This proves (\ref{eq:cov_birth_C_E_x}) and completes the proof of
Proposition \ref{prop:mean_cov_birth}.

\section{\label{sec:Appendix_C}}

In this appendix, we prove Lemma \ref{lem:Steady_state_appearance}.
We first prove the mean and then the covariance matrix.

\subsection{Mean}

Substituting (\ref{eq:mean_steady}) into (\ref{eq:E_x_t}) yields
\begin{align}
\mathrm{E}\left[x_{k}\left|t\right.\right] & =-\exp\left(At\right)A^{-1}u+\int_{0}^{t}\exp\left(A\tau\right)d\tau u.
\end{align}
As $A$ is invertible, we have that 
\begin{align}
\mathrm{E}\left[x_{k}\left|t\right.\right] & =-\exp\left(At\right)A^{-1}u+\left[\exp\left(At\right)-I\right]A^{-1}u\nonumber \\
 & =-A^{-1}u.\nonumber \\
 & =x_{\infty}.
\end{align}
Then, it is direct to obtain that
\begin{align}
\overline{x}_{b,k} & =\mathrm{E}\left[\mathrm{E}\left[x_{k}\left|t\right.\right]\right]\nonumber \\
 & =x_{\infty}.
\end{align}
This proves the result for the mean.

\subsection{Covariance matrix}

Substituting (\ref{eq:cov_steady}) into (\ref{eq:C_x_t})

\begin{align}
\mathrm{C}\left[x_{k}\left|t\right.\right] & =\exp\left(At\right)P_{\infty}\exp\left(A^{T}t\right)\nonumber \\
 & \quad+\int_{0}^{t}\exp\left(A\tau\right)LQ_{\beta}L^{T}\exp\left(A^{T}\tau\right)d\tau\nonumber \\
 & =\exp\left(At\right)\left[\int_{0}^{\infty}\exp\left(A\tau\right)LQ_{\beta}L^{T}\exp\left(A^{T}\tau\right)d\tau\right]\nonumber \\
 & \quad\times\exp\left(A^{T}t\right)\nonumber \\
 & \quad+\int_{0}^{t}\exp\left(A\tau\right)LQ_{\beta}L^{T}\exp\left(A^{T}\tau\right)d\tau\nonumber \\
 & =\int_{0}^{\infty}\exp\left(A\left(t+\tau\right)\right)LQ_{\beta}L^{T}\exp\left(A^{T}\left(t+\tau\right)\right)d\tau\nonumber \\
 & \quad+\int_{0}^{t}\exp\left(A\tau\right)LQ_{\beta}L^{T}\exp\left(A^{T}\tau\right)d\tau.
\end{align}
Making the change of variable $l=\tau+t$ in the first integral, we
obtain
\begin{align}
\mathrm{C}\left[x_{k}\left|t\right.\right] & =\int_{t}^{\infty}\exp\left(Al\right)LQ_{\beta}L^{T}\exp\left(A^{T}l\right)dl\nonumber \\
 & \quad+\int_{0}^{t}\exp\left(A\tau\right)LQ_{\beta}L^{T}\exp\left(A^{T}\tau\right)d\tau\nonumber \\
 & =\int_{0}^{\infty}\exp\left(A\tau\right)LQ_{\beta}L^{T}\exp\left(A^{T}\tau\right)d\tau\nonumber \\
 & =P_{\infty}.
\end{align}
Then, it is direct to obtain that
\begin{align}
P_{b,k} & =\mathrm{C}\left[\mathrm{E}\left[x_{k}\left|t\right.\right]\right]+\mathrm{E}\left[\mathrm{C}\left[x_{k}\left|t\right.\right]\right]\nonumber \\
 & =0+P_{\infty}.
\end{align}
This finishes the proof of Lemma \ref{lem:Steady_state_appearance}.

\section{\label{sec:Appendix_D}}

In this appendix, we prove Lemma \ref{lem:Steady_state_lambda}. We
first prove (\ref{eq:predicted_lambda_steady}) and then (\ref{eq:updated_lambda_steady}).

\subsection{Proof of (\ref{eq:predicted_lambda_steady})}

Using (\ref{eq:mean_intensity_prediction}) and (\ref{eq:mean_intensity_update})
and the conditions in Lemma \ref{lem:Steady_state_lambda}, the predicted
mean number of undetected targets is
\begin{align}
\overline{\lambda}_{k|k-1} & =p^{S}\left(1-p^{D}\right)\overline{\lambda}_{k-1|k-2}+\overline{\lambda}^{B}.
\end{align}
In the steady-state solution, $\overline{\lambda}_{k|k-1}=\overline{\lambda}_{k-1|k-2}=\overline{\lambda}_{k|k-1}^{\infty}$,
which yields
\begin{align}
\overline{\lambda}_{k|k-1}^{\infty} & =p^{S}\left(1-p^{D}\right)\overline{\lambda}_{k|k-1}^{\infty}+\overline{\lambda}^{B},
\end{align}
\begin{align}
\left[1-p^{S}\left(1-p^{D}\right)\right]\overline{\lambda}_{k|k-1}^{\infty} & =\overline{\lambda}^{B}.\label{eq:lambda_steady_append}
\end{align}
This equation can be solved if $\left[1-p^{S}\left(1-p^{D}\right)\right]\neq0$,
which is ensured if $p^{S}\neq1$ or $p^{D}\neq0$. As $\mu>0$, it
is met that $p^{S}<1$, see (\ref{eq:Probability_survival}), so (\ref{eq:lambda_steady_append})
always has a solution. The solution of this equation proves (\ref{eq:predicted_lambda_steady}).
It should be noted that if we had that $p^{S}=1$ and $p^{D}=0$,
there is no steady state solution as $\overline{\lambda}_{k|k-1}$
increases with time.

\subsection{Proof of (\ref{eq:updated_lambda_steady})}

Using (\ref{eq:mean_intensity_prediction}) and (\ref{eq:mean_intensity_update})
and the conditions in Lemma \ref{lem:Steady_state_lambda}, the updated
mean number of undetected targets is
\begin{align}
\overline{\lambda}_{k|k} & =\left(1-p^{D}\right)\left(p^{S}\overline{\lambda}_{k-1|k-1}+\overline{\lambda}^{B}\right).
\end{align}
In the steady-state solution, $\overline{\lambda}_{k|k}=\overline{\lambda}_{k-1|k-1}=\overline{\lambda}_{k|k}^{\infty}$,
which yields
\begin{align}
\overline{\lambda}_{k|k}^{\infty} & =\left(1-p^{D}\right)\left(p^{S}\overline{\lambda}_{k|k}^{\infty}+\overline{\lambda}^{B}\right).
\end{align}
Solving this equation, which is possible with $p^{S}\neq1$ or $p^{D}\neq0$,
proves (\ref{eq:updated_lambda_steady}).

\section{\label{sec:Appendix_E}}

In this appendix, we prove Proposition \ref{prop:Birth_nonlinear}.
We first compute the mean and then the covariance matrix.

\subsection{Mean at the time of birth}

Using (\ref{eq:conditional_mean_cov_birth_nonlinear}) and (\ref{eq:mean_birth_prev}),
the mean at the time of birth is
\begin{align}
\overline{x}_{b,k} & =\mathrm{E}\left[\mathrm{E}\left[x_{k}\left|t\right.\right]\right]\nonumber \\
 & =\frac{\mu}{1-e^{-\mu\Delta t_{k}}}\int_{0}^{\Delta t_{k}}m\left(t\right)e^{-\mu t}dt.
\end{align}
It is direct to check that
\begin{align*}
\overline{x}_{b,k} & =\frac{\mu}{1-e^{-\mu\Delta t_{k}}}\overline{x}\left(\Delta t_{k}\right)
\end{align*}
where $\overline{x}\left(\Delta t_{k}\right)$ is the solution of
the ODE
\begin{align*}
\frac{d\overline{x}}{dt} & =me^{-\mu t}
\end{align*}
with initial condition $\overline{x}\left(0\right)=0$ at time $t=\Delta t_{k}$. 

\subsection{Covariance at the time of birth}

Following (\ref{eq:mean_birth_prev}), the covariance matrix at the
time of birth can be written as
\begin{align}
P_{b,k} & =P_{b,k}^{1}-\overline{x}_{b,k}\overline{x}_{b,k}^{T}
\end{align}
where
\begin{align}
P_{b,k}^{1} & =\mathrm{E}\left[\mathrm{C}\left[x_{k}\left|t\right.\right]+\mathrm{E}\left[x_{k}\left|t\right.\right]\mathrm{E}\left[x_{k}\left|t\right.\right]^{T}\right]\nonumber \\
 & =\frac{\mu}{1-e^{-\mu\Delta t_{k}}}\int\left[P\left(t\right)+m\left(t\right)m^{T}\left(t\right)\right]e^{-\mu t}dt.
\end{align}
It is direct to check that this integral can be computed as 
\begin{align}
P_{b,k}^{1} & =\frac{\mu}{1-e^{-\mu\Delta t_{k}}}\Sigma\left(\Delta t_{k}\right)
\end{align}
where $\Sigma\left(\Delta t_{k}\right)$ is the solution of the ODE
\begin{align}
\frac{d\Sigma}{dt} & =\left[P+mm^{T}\right]e^{-\mu t}
\end{align}
with initial condition $\Sigma\left(0\right)=0$ from $t=0$ to $t=\Delta t_{k}$.
Then, to calculate $\overline{x}\left(\Delta t_{k}\right)$ and $\Sigma\left(\Delta t_{k}\right)$,
we must simultaneously solve the ODEs in Proposition \ref{prop:Birth_nonlinear},
which also account for the evolution of $m\left(t\right)$ and $P\left(t\right)$
with time.
\end{document}